\documentclass[nohyperref]{article}

\usepackage{caption}
\usepackage{subcaption}
\usepackage{microtype}
\usepackage{graphicx}
\usepackage{booktabs} 

\usepackage{hyperref}

\usepackage[accepted]{icml2022}

\usepackage{amsmath}
\usepackage{amssymb}
\usepackage{mathtools}
\usepackage{amsthm}

\usepackage[capitalize,noabbrev]{cleveref}

\theoremstyle{plain}

\theoremstyle{definition}

\theoremstyle{remark}

\usepackage[textsize=tiny]{todonotes}

\icmltitlerunning{Success of Uncertainty-Aware Deep Models Depends on Data Manifold Geometry}

\begin{document}
\twocolumn[
\icmltitle{Success of Uncertainty-Aware Deep Models Depends on Data Manifold Geometry}



\icmlsetsymbol{equal}{*}

\begin{icmlauthorlist}
\icmlauthor{Mark Penrod}{equal,harvard}
\icmlauthor{Harrison Termotto}{equal,harvard}
\icmlauthor{Varshini Reddy}{equal,harvard}
\icmlauthor{Jiayu Yao}{seas}
\icmlauthor{Finale Doshi-Velez}{seas}
\icmlauthor{Weiwei Pan}{harvard,seas}
\end{icmlauthorlist}

\icmlaffiliation{harvard}{IACS, Harvard University, Cambridge, MA}
\icmlaffiliation{seas}{SEAS, Harvard University, Cambridge, MA}

\icmlcorrespondingauthor{Mark Penrod}{mpenrod@g.harvard.edu}
\icmlcorrespondingauthor{Harrison Termotto}{harrisontermotto@g.harvard.edu}
\icmlcorrespondingauthor{Varshini Reddy}{varshinibogolu@g.harvard.edu}

\icmlkeywords{Machine Learning, Uncertainty, Deep Learning, Neural Network, ICML}

\vskip 0.3in
]



\printAffiliationsAndNotice{\icmlEqualContribution} 

\begin{abstract}
For responsible decision making in safety-critical settings,
machine learning models must effectively detect and process edge-case data. Although existing works show that predictive uncertainty is useful for these tasks, it is not evident from literature which uncertainty-aware models are best suited for a given dataset.
Thus, we compare six uncertainty-aware deep learning models on a set of edge-case tasks: robustness to adversarial attacks as well as out-of-distribution and adversarial detection. We find that the geometry of the data sub-manifold is an important factor in determining the success of various models. Our finding suggests an interesting direction in the study of uncertainty-aware deep learning models.

\end{abstract}

\section{Introduction}
\label{intro}

Responsible development and deployment of machine learning models in high-stakes real-life applications often require these models to have guarantees of desirable behaviors on edge-cases. In literature, the ability to handle edge-cases is often formalized as model's awareness of out-of distribution (OoD) and adversarial data, and its adversarial robustness. Although many recent works have developed models with those desired properties \cite{epiunc, bdl_diabetic_retinopathy, thresholding_mi_adv}, it is not clear from literature which subset of existing uncertainty-aware deep models is generally well suited for a given dataset. 

In this work, we evaluate the usefulness of a set of commonly used Bayesian and non-Bayesian uncertainty-aware deep models for (1) adversarial robustness, and (2) detection of adversarial and OoD data. Furthermore, we study to what extent commonly used metrics of uncertainty (e.g. predictive entropy) can be used as proxies for model performance on these tasks. Our key finding is that the effectiveness of uncertainty-aware models, as well as the usefulness of uncertainty metrics, varies depending on the geometry of the data. Specifically, we find that: (1) models that succeed on our tasks when the data sub-manifold is low dimensional (compared to the input space) may fail when the data manifold has co-dimension zero, and vice versa; (2) the ability of different uncertainty metrics to capture various aspects of the model's predictive uncertainty also depends on the dimension of the data sub-manifold. To our knowledge, we are the first to study how the geometry of data sub-manifolds effects the usefulness of uncertainty-aware deep models and uncertainty metrics. 

\begin{figure*}[ht]
    \centering
    \begin{subfigure}[b]{0.6\columnwidth}
         \centering
         \includegraphics[width=\textwidth]{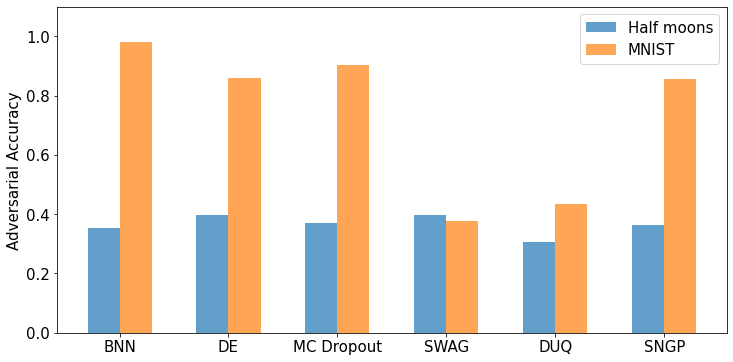}
         \caption{Adversarial Accuracy}
     \end{subfigure}
     \hfill
    \begin{subfigure}[b]{0.6\columnwidth}
         \centering
         \includegraphics[width=\textwidth]{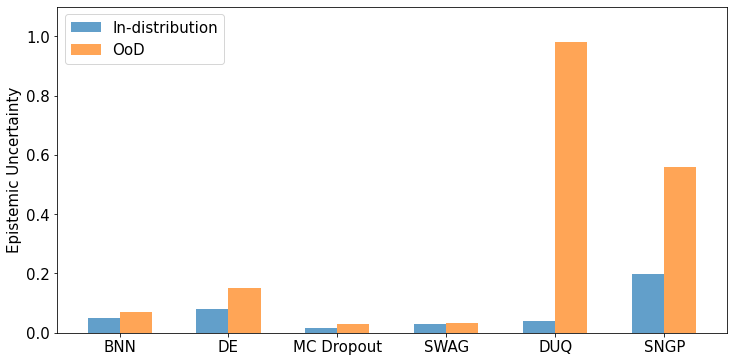}
         \caption{Epistemic Uncertainty (half moons)}
     \end{subfigure}
     \hfill
    \begin{subfigure}[b]{0.6\columnwidth}
         \centering
         \includegraphics[width=\textwidth]{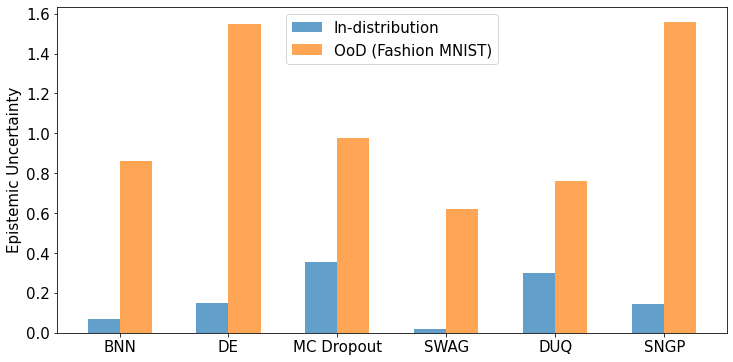}
         \caption{Epistemic Uncertainty (MNIST)}
     \end{subfigure}
     \hfill
    \caption{Model performance by dataset. (a) Certain models show clear increase in robustness to adversarial attacks in the MNIST setting over half moons. (b) For half moons data, ability to detect OoD points via uncertainty is poor for all models except single-pass. (c) In contrast, all models produce high uncertainty in for Fashion MNIST setting, relative to uncertainty on MNIST}
    \label{adv-accuracies}
\end{figure*}
\section{Related Work}
\label{relatedwork}

Past work shows that uncertainty-aware deep learning models can be used to detect out-of-distribution data \cite{single-pass-ood, sngp, origde}. Furthermore, existing works show that these models are more robust to adversarial attacks \cite{deepgp_robust, ensemblesrobust, bnns_robust, laplace_oodadv, adv_samples_from_artifacts} and that methods for detecting OoD data can also be utilized for detecting adversarial data \cite{understanding_measures, thresholding_mi_adv, carlini_ten_methods_adv_detection}. 
However, there is very little work on how to select amongst  available uncertainty-aware deep models for a given general dataset. While previous works compare various types of models, such comparisons often focus on bench-marking model performance on a range of specific datasets \cite{unccomp, unccomp2, koh2021wilds}. 
However, such approaches do not always characterize the differences in model performance in terms of the intrinsic properties of the datasets themselves, making it potentially difficult to generalize insights to new settings. In contrast, we evaluate differences in model performance in relation to the inherent geometry of data sub-manifolds. 

At the same time, model predictive uncertainty is often summarized by a choice of uncertainty metric (e.g. predictive entropy) that acts as a proxy for the intended task performance (e.g. adversarial robustness, detection of adversarial and OoD data). For example, high uncertainty as measured by entropy can be used to flag edge-cases for human review  \cite{bdl_diabetic_retinopathy}. However, there are many uncertainty metrics one can consider but few works that compare them as proxies for our intended task performance. In this paper, we compare several commonly used uncertainty metrics in their ability to indicate model performance on adversarial robustness as well as detection of adversarial and OoD data. Here too, we relate the differences in the usefulness of uncertainty metrics to the geometry of the data sub-manifolds.

\section{Background}
\paragraph{Adversarial Data}
\label{adv-attacks}
Let $x_n \in \mathbb{R}^D$ be a input vector and $\mathcal{F}$ be a classification model with predicted class label $y_n\in N$. 
Of the many types of adversarial attacks, we consider adding a small perturbation $\delta \in \mathbb{R}^D$ to the input $x_n$, which fools the model (i.e. $\mathcal{F}(x_n + \delta) \ne y_n$) but does not fool a human. A model is robust to adversaries if it is able to correctly classify inputs modified by adversarial perturbations. 

For our experiments, we consider a black-box attack that can only query the target model predictions but not the model parameters, which is a common practical constraint. Specifically, we take inspiration from previous work in black-box attacks \cite{bboxattack2, bboxattack} and train a deterministic proxy model that approximates the target model's decision boundary. We then execute an L2-PGD attack on the trained proxy to generate adversarial examples (\hyperref[adv-appendix]{Appendix D}).
\paragraph{Out-of-Distribution Data} Ideally, during test time, machine learning models will encounter data drawn from the same data generative process as the training data. However, in certain situations, such as the presence of anomalous examples or covariate shift, the new incoming data may not resemble the training data. We refer to such data as out-of-distribution (OoD) samples and the task of distinguishing among in-distribution and OoD data as OoD detection. 
\vspace{-0.3cm}
\section{Experiment Setup}
\label{methods}
\paragraph{Datasets} We consider three datasets: (1) the half moons dataset for binary classification, (2) MNIST for multi-class image classification, (3) a binary classification dataset wherein data clusters primarily lie on a 1-dimensional line within 2-dimensional space to mimic the natural image manifold. (1) and (2) are commonly used benchmark datasets, (3) is a pedagogical dataset designed to demonstrate a low dimensional data sub-manifold structure.

For datasets (1) and (2), adversarial attacks are generated per model by adding adversarial perturbation to the inputs determined via a gradient-based attack to the proxy. For dataset (3), separate on and off-manifold adversaries are generated by perturbing the inputs by $\delta \sim \mathcal{N}(\pm\epsilon, 0.5)$, where $\epsilon$ is a strength parameter (\hyperref[adv-appendix]{Appendix D}).

For datasets (1) and (3), OoD clusters are hand placed in several distant positions in the feature space. For dataset (2), we use Fashion MNIST as OoD examples.

\paragraph{Models}

We consider a set of commonly used uncertainty-aware deep models. We study two Bayesian approaches: \textbf{Bayesian Neural Networks (BNNs)}, in which we place a prior on the neural network weights and infer the posterior via Hamiltonian Monte Carlo; and \textbf{Stochastic Weighted Average Gaussian (SWAG)} \cite{swag}, which approximates the posterior of BNNs by parametrizing a Gaussian distribution around the iterates of averaged SGD solutions. We also study two frequentist approaches: \textbf{Deep Ensembles}, which aggregates a collection of independently trained neural networks (we note that although several methods exist for producing variance in ensembles \cite{hyperparamde, repuslivede}, we only consider variance through random initialization of model weights \cite{origde}); and \textbf{Monte Carlo (MC) Dropout} \cite{mc_dropout}, which provides a distribution over predictions via a sequence of dropout-enabled forward passes through a pre-trained network. Last, we employ two single-pass, deterministic models: \textbf{Deterministic Uncertainty Quantification (DUQ)} \cite{duq} and \textbf{Spectral-Normalized Neural Gaussian Process (SNGP)} \cite{sngp}. Both models encode predictive uncertainty via forms of distance-awareness \cite{duq, sngp}.

\paragraph{Uncertainty Metrics}

We measure two types of uncertainty for all models: \textit{epistemic uncertainty}, quantifying the uncertainty of the model due to insufficiency of observations, and \textit{aleatoric uncertainty}, quantifying the uncertainty due to the inherent stochasticity in the data generating process. 

For ensemble-based models, we quantify \textit{aleatoric entropy} as the average entropy of the probability vectors output by each component of the ensemble. We quantify \textit{epistemic entropy} under two frameworks with each distilling the degree to which the ensemble components disagree. First, we measure the entropy of the average predicted probability across models \cite{epiunc}. We note that this method may not always correctly identify instances in which the components disagree. To address this, we propose an alternative method for measuring epistemic uncertainty wherein we calculate Kullback-Leibler divergence for all combinations of ensemble predictions and average to produce a single value per datum (\textit{KL uncertainty}).

Each single-pass model requires its own method for uncertainty quantification. For DUQ, we use the distance from the assigned class centroid as epistemic uncertainty \cite{duq} and entropy of the output correlations, transformed to valid probabilities, as aleatoric uncertainty. For SNGP, we draw  samples from the predictive distribution, $\mathcal{N}(\text{logit}(x_n), \text{var}(x_n))$, in order to form an ensemble of predictions for each data point. These predictions are used to compute aleatoric and epistemic uncertainties in the same manner as the other ensemble models.


\begin{figure}[h]
    \centering
    \includegraphics[width=0.34\columnwidth]{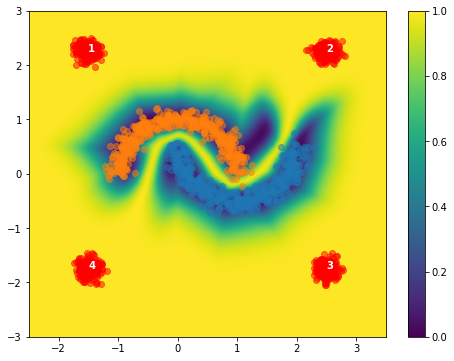}
    \includegraphics[width=0.34\columnwidth]{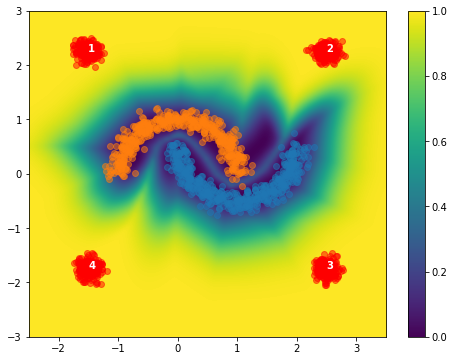}
    \includegraphics[width=0.32\columnwidth]{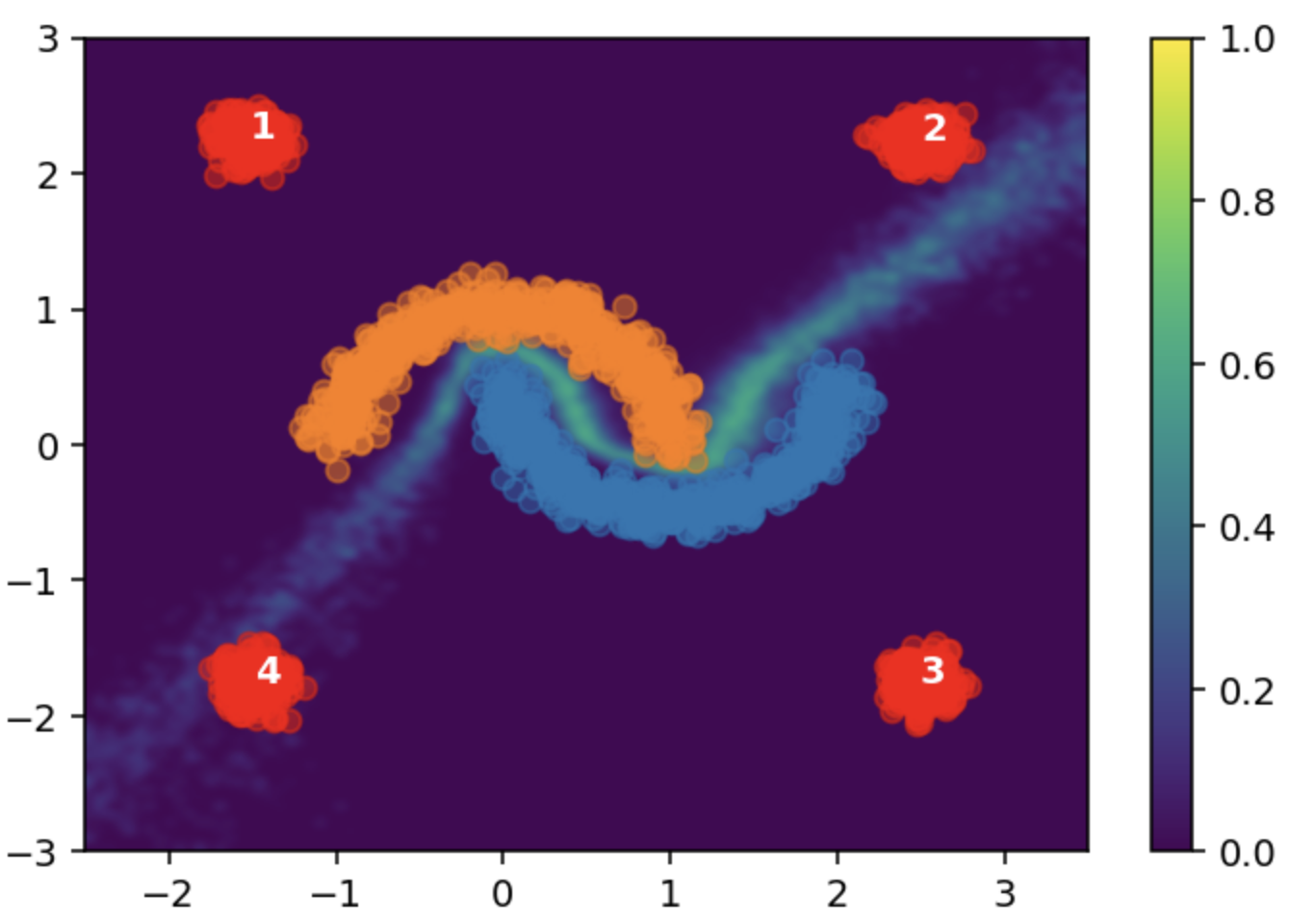}    \includegraphics[width=0.32\columnwidth]{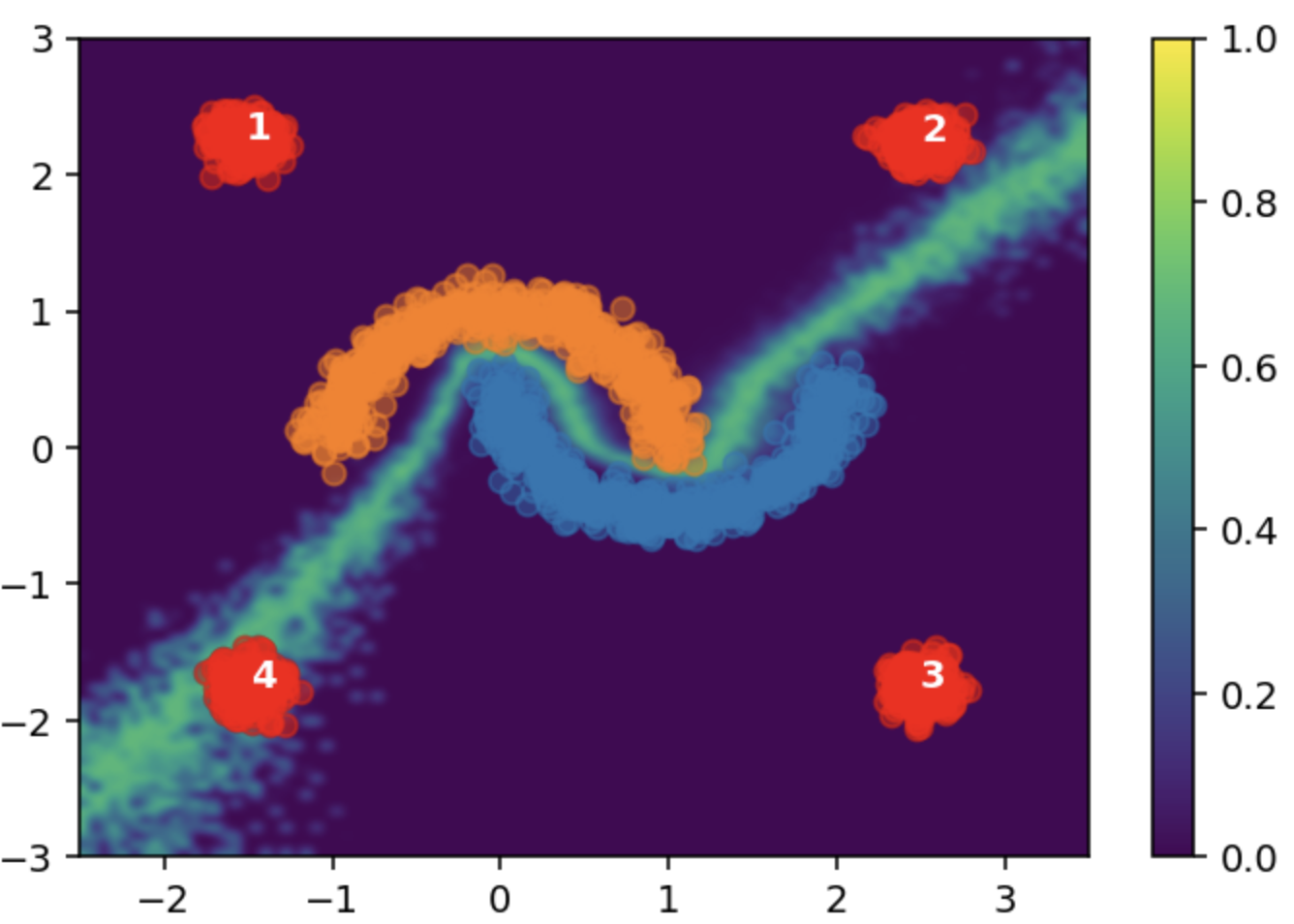}
    \includegraphics[width=0.32\columnwidth]{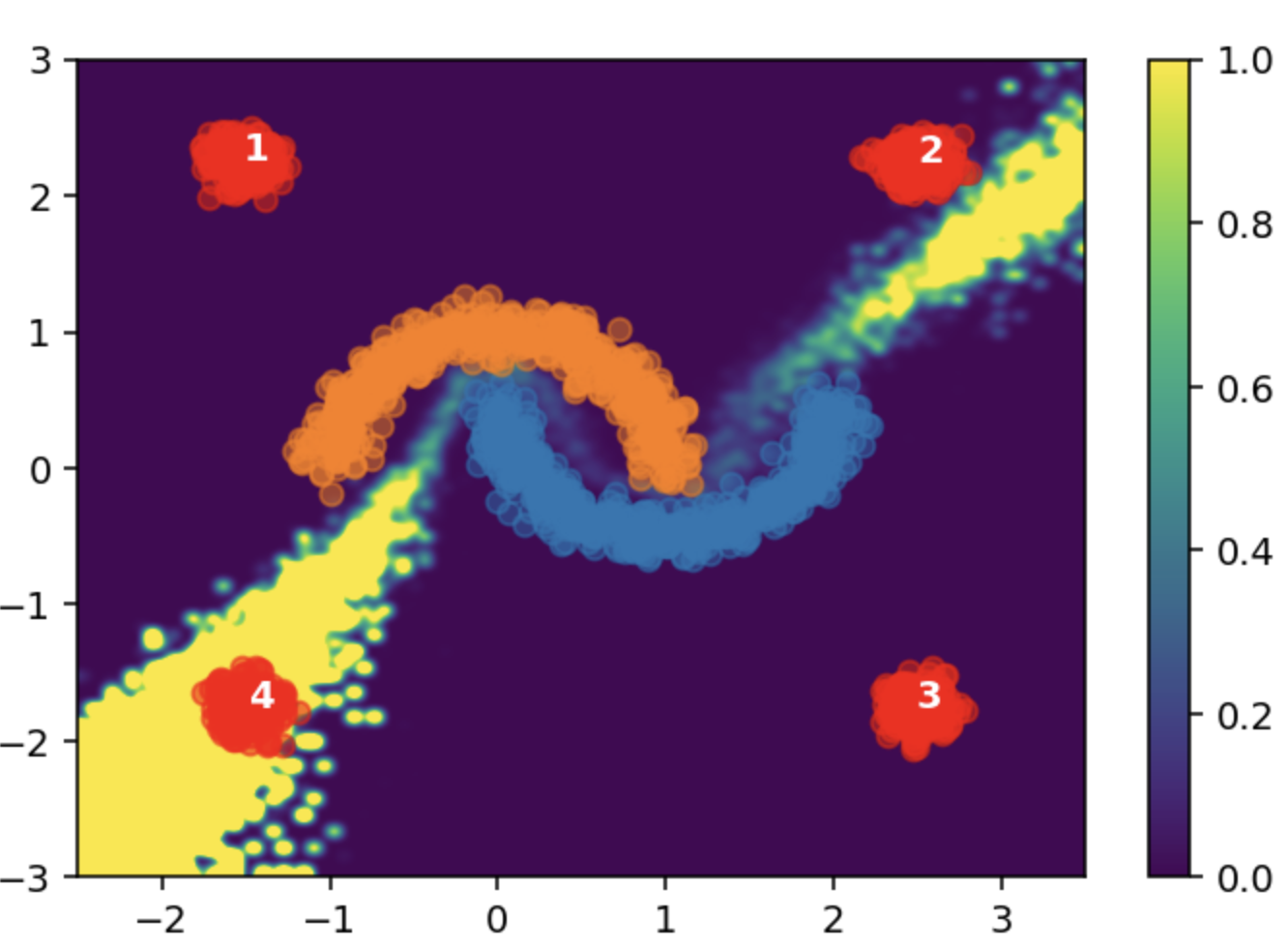}
    \caption{Uncertainties on half moons data for DUQ (top) and MC Dropout (bottom). For DUQ, uncertainty is divided into aleatoric entropy (left) and epistemic (right). MC Dropout uncertainty is divided into aleatoric entropy (left), epistemic entropy (middle), and epistemic KL-divergence (right). Out-of-distribution points are designated by red clusters.}
    \label{hm-uncs}
\end{figure}

\section{Results \& Analyses}
\label{results}
\paragraph{No single uncertainty-aware model performs best across all tasks.} On the half moons dataset, all models show a relatively consistent robustness to adversarial attacks. However, the single-pass models, i.e. DUQ and SNGP, are much more sensitive to out-of-distribution data (\hyperref[adv-table]{Table 1}, \hyperref[adv-accuracies]{Figure 1}), as both models produce highly-curved decision boundaries and project uncertainty into all areas of the feature space outside the data-rich region (\hyperref[appendix-halfmoons-uncertainties]{Appendix E}). All other models produce uncertainty that increases outside of the data-region, but cling to a simpler decision boundary. Thus, the single-pass models are more sensitive to OoD data regardless of the data positions in feature space.

However, this pattern does not hold on natural image datasets (MNIST). First, some models demonstrate substantial adversarial robustness where others are more vulnerable. Next, all models now show high levels of uncertainty on OoD data. Unlike the half moons example, the discrepancy in performance does not follow any clear delineation along model classes (\hyperref[adv-accuracies]{Figure 1}). 

So what can explain the discrepancy between the trends in model performance on the two datasets? In the following, we characterize these discrepancies in terms of the intrinsic data geometries.


\begin{figure*}[h]
\begin{center}
\includegraphics[width=4cm, height=2.7cm]{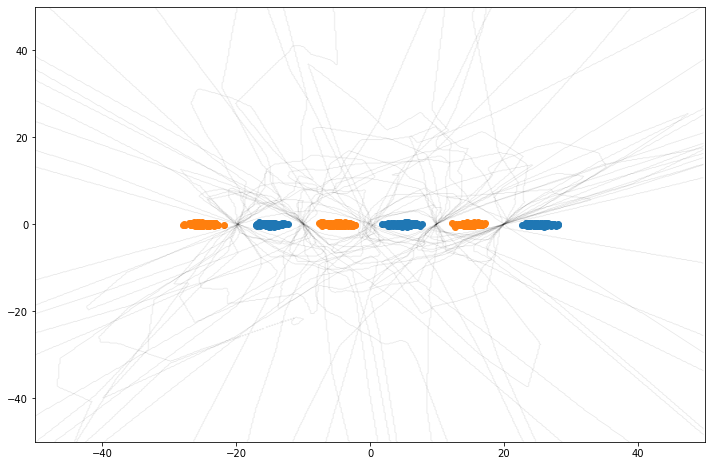}
\includegraphics[width=4.1cm, height=2.7cm]{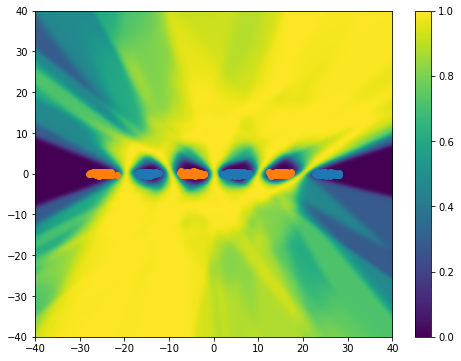}
\includegraphics[width=4cm, height=2.7cm]{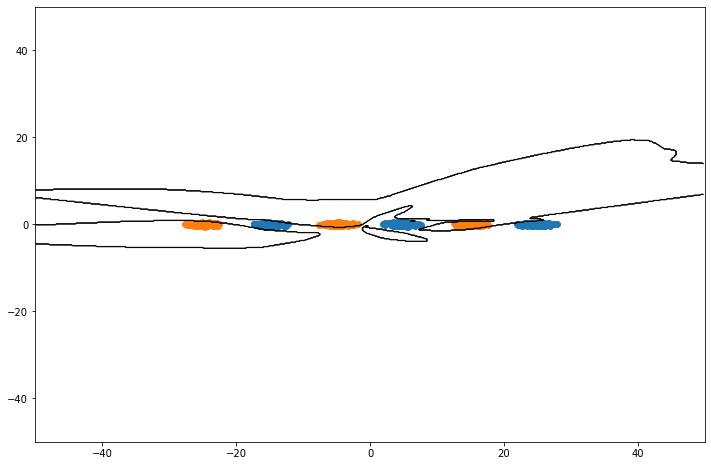}
\includegraphics[width=4.1cm, height=2.7cm]{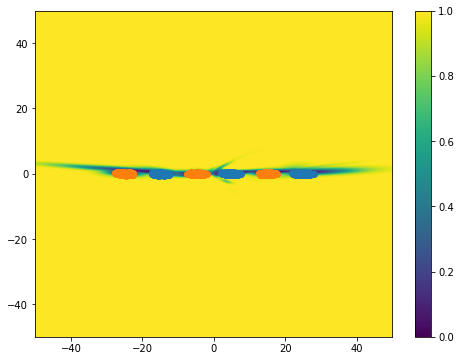}
\end{center}
\caption{Decision boundaries and epistemic uncertainties for Deep Ensemble (left) and DUQ (right) in toy manifold. Deep ensembles produce a variety of decision boundaries that extend into off manifold space. DUQ by definition draws a single decision boundary and the model confidence tightly hugs the data.}
\label{toymani_db_unc}
\end{figure*}

\textbf{Model performance differences depend on the geometry of the data sub-manifolds.}\quad There is a consensus in literature that natural images lie on or near low-dimensional manifolds relative to the high-dimensional pixel space \cite{imagemanifold1,imagemanifold2}. In contrast, the data manifold of half moons is full-dimensional, as data variance is significant in all input dimensions. To investigate the relationship between geometry of the data manifold and performance on detection and robustness tasks, we consider a simple dataset composed of clusters that lie on a simple, 1D linear manifold within a 2D space. 

First, \hyperref[toymani_db_unc]{Figure 3} provides insight into the remediated discrepancy in sensitivity to OoD data. In this toy setting, deep ensembles – and to a lesser degree SWAG, BNN, and MC Dropout – now generate diverse decision boundaries by exploiting the freedom in higher dimensions and thus encode uncertainty omnidirectionally in off-manifold space (\hyperref[appendix-toy-manifold-uncertainties]{Appendix F}). This behavior mimics that of single-pass models (i.e. DUQ and SNGP) on both half moons and toy manifold experiments. In contrast to some previous works \cite{ensembles_skep}, this observation highlights how ensemble diversity is critical for accurate and useful uncertainty. 

Next, when given off-manifold adversaries of increasing strength, we see differences in robustness similar to MNIST. Notably, all models show similar vulnerability to on-manifold adversarial perturbations (\hyperref[toy-mani-robustness]{Figure 6}). Taking deep ensembles and DUQ as examples, we see that deep ensembles generate varied decision boundaries due to the flexibility in off-manifold space, meaning that adversarial attacks may cross one decision boundary, but will remain correctly assigned by others. As such, the boundaries of ensembles tend to be adversarially robust, a feature DUQ naturally lacks due to its single, deterministic boundary. 

Results of this experiment highlight another way that the geometry of the data sub-manifold affects model performance: geometry of the loss surface. The clustered data on the 1-D manifold allows many probable decision boundaries and, as such, is a highly non-convex loss surface. As a result, model performance for SWAG, BNN, and MC Dropout can vary greatly by random restart (\hyperref[mc-random-restarts]{Figure 4}).\footnote{Each of these models train a single network and in distinct manners explore the area surrounding the given local minimum identified by training. This feature results in variable behavior for highly modal loss spaces.} With a smoother loss surface, we expect and observe more stable behavior (as in the case of MNIST experiments).

\textbf{Effectiveness of uncertainty metrics as a proxy of the model's ability to handle edge-cases depends on data sub-manifold geometry.}\quad Just as there are numerous deep models that produce predictive uncertainty, there are many metrics for quantifying the predictive uncertainty of these models. We find that the degree to which each metric can be used to gauge model performance depends on the geometry of the data sub-manifold.

In \hyperref[hm-uncs]{Figure 2}, we can see the uncertainty in the feature space under three metrics: mean predictive entropy (aleatoric), entropy of predictive mean (epistemic), and KL uncertainty between model predictions (epistemic). In both half moons and the toy manifold, as expected, we see that aleatoric entropy prioritizes the areas immediately between clusters, since misclassification due to noise is most likely where the clusters overlap, and KL focuses on regions away from the data, as incongruence among model predictions is most likely where training data is scarce. Epistemic entropy, however, spans both of these regions as it also increases in areas with low density of training data, but is left-bounded by aleatoric entropy.

These patterns have important implications in terms of adversarial and OoD awareness. For ensemble-based models on full dimensional data, aleatoric entropy is more sensitive to adversaries, since adversaries in this case lie close to the decision boundary. KL is sensitive to OoD data. This is because KL measures pairwise disagreement between the ensemble of predictions and, in this case, OoD data lie in regions where there are fewer data observations and that admits diverse decision boundaries. Finally, epistemic entropy serves as a catch-all for both since it quantifies both uncertainty over the predicted label as well as disagreements in the ensemble.
However, these patterns change when the data sub-manifold is low-dimensional. In this setting, KL and epistemic entropy are generally more effective for detecting adversaries and OoD, since when the data sub-manifold is low-dimensional both adversarial examples and OoD are likely to fall off-manifold; aleatoric entropy will quantify uncertainty for on-manifold data, since it is measuring where the classes overlap. %


\section{Conclusion \& Discussion}
\label{conclusion}
In this work, we evaluate a set of deep uncertainty quantification models on their robustness to and ability to detect adversarial and out-of-distribution examples. We also study the extent to which commonly used uncertainty metrics are good proxies for model performance on these tasks. We show that the success of uncertainty-aware models and uncertainty metrics depends on the data geometry. Specifically, we observe that performance discrepancies can be traced to the presence of a low-dimensional data manifold. 

Insights from these findings have important implications for the application of uncertainty-aware deep learning methods. 
Namely, for datasets that are not reducible to a manifold, such as half moons,  methods in the vein of DUQ and SNGP are well-suited for OoD and adversarial detection. 
If using any other tested models, KL-based measurements are better suited for OoD detection, while aleatoric entropy will allow for specific detection of adversaries. 
However, should the data lie on a low-dimensional manifold, as is the case for natural images, model choice will have a strong impact on adversarial robustness, but the ability to detect OoD or adversarial data via uncertainty is generally assured. Additionally, aleatoric entropy in this case is more appropriate for on-manifold data detection (e.g. OoD images), while the KL-based metric is better at detecting data that projects off-manifold (e.g. adversarial images).

The choices of model and uncertainty metric interact with data geometry and task-related desiderata in complex and potentially non-intuitive ways. As such, this work emphasizes the importance of a careful analysis of those factors when deploying uncertainty-aware deep learning models. 
\section{Acknowledgement}
JY and FDV acknowledge the support from NSF IIS-2007076.
\bibliography{main}

\begin{thebibliography}{27}
\providecommand{\natexlab}[1]{#1}
\providecommand{\url}[1]{\texttt{#1}}
\expandafter\ifx\csname urlstyle\endcsname\relax
  \providecommand{\doi}[1]{doi: #1}\else
  \providecommand{\doi}{doi: \begingroup \urlstyle{rm}\Url}\fi

\bibitem[Abe et~al.(2022)Abe, Buchanan, Pleiss, Zemel, and
  Cunningham]{ensembles_skep}
Abe, T., Buchanan, E.~K., Pleiss, G., Zemel, R., and Cunningham, J.~P.
\newblock Deep ensembles work, but are they necessary?, 2022.

\bibitem[Arnez et~al.(2020)Arnez, Espinoza, Radermacher, and Terrier]{unccomp2}
Arnez, F., Espinoza, H., Radermacher, A., and Terrier, F.
\newblock A comparison of uncertainty estimation approaches in deep learning
  components for autonomous vehicle applications.
\newblock \emph{CoRR}, abs/2006.15172, 2020.

\bibitem[Bradshaw et~al.(2017)Bradshaw, de~G.~Matthews, and
  Ghahramani]{deepgp_robust}
Bradshaw, J.~F., de~G.~Matthews, A.~G., and Ghahramani, Z.
\newblock Adversarial examples, uncertainty, and transfer testing robustness in
  gaussian process hybrid deep networks.
\newblock \emph{arXiv: Machine Learning}, 2017.

\bibitem[Carbone et~al.(2020)Carbone, Wicker, Laurenti, Patane, Bortolussi, and
  Sanguinetti]{bnns_robust}
Carbone, G., Wicker, M., Laurenti, L., Patane, A., Bortolussi, L., and
  Sanguinetti, G.
\newblock Robustness of bayesian neural networks to gradient-based attacks,
  2020.

\bibitem[Carlini \& Wagner(2017)Carlini and
  Wagner]{carlini_ten_methods_adv_detection}
Carlini, N. and Wagner, D.
\newblock Adversarial examples are not easily detected: Bypassing ten detection
  methods, 2017.

\bibitem[D'Angelo \& Fortuin(2021)D'Angelo and Fortuin]{repuslivede}
D'Angelo, F. and Fortuin, V.
\newblock Repulsive deep ensembles are bayesian, 2021.

\bibitem[Feinman et~al.(2017)Feinman, Curtin, Shintre, and
  Gardner]{adv_samples_from_artifacts}
Feinman, R., Curtin, R.~R., Shintre, S., and Gardner, A.~B.
\newblock Detecting adversarial samples from artifacts, 2017.

\bibitem[Filos et~al.(2019)Filos, Farquhar, Gomez, Rudner, Kenton, Smith,
  Alizadeh, de~Kroon, and Gal]{bdl_diabetic_retinopathy}
Filos, A., Farquhar, S., Gomez, A.~N., Rudner, T. G.~J., Kenton, Z., Smith, L.,
  Alizadeh, M., de~Kroon, A., and Gal, Y.
\newblock A systematic comparison of bayesian deep learning robustness in
  diabetic retinopathy tasks, 2019.

\bibitem[Gal \& Ghahramani(2015)Gal and Ghahramani]{mc_dropout}
Gal, Y. and Ghahramani, Z.
\newblock Dropout as a bayesian approximation: Representing model uncertainty
  in deep learning, 2015.

\bibitem[Kendall \& Gal(2017)Kendall and Gal]{epiunc}
Kendall, A. and Gal, Y.
\newblock What uncertainties do we need in bayesian deep learning for computer
  vision?, 2017.

\bibitem[Koh et~al.(2021)Koh, Sagawa, Marklund, Xie, Zhang, Balsubramani, Hu,
  Yasunaga, Phillips, Gao, et~al.]{koh2021wilds}
Koh, P.~W., Sagawa, S., Marklund, H., Xie, S.~M., Zhang, M., Balsubramani, A.,
  Hu, W., Yasunaga, M., Phillips, R.~L., Gao, I., et~al.
\newblock Wilds: A benchmark of in-the-wild distribution shifts.
\newblock In \emph{International Conference on Machine Learning}, pp.\
  5637--5664. PMLR, 2021.

\bibitem[Lakshminarayanan et~al.(2017)Lakshminarayanan, Pritzel, and
  Blundell]{origde}
Lakshminarayanan, B., Pritzel, A., and Blundell, C.
\newblock Simple and scalable predictive uncertainty estimation using deep
  ensembles.
\newblock In \emph{Advances in Neural Information Processing Systems},
  volume~30. Curran Associates, Inc., 2017.

\bibitem[Liu et~al.(2020)Liu, Lin, Padhy, Tran, Bedrax{-}Weiss, and
  Lakshminarayanan]{sngp}
Liu, J.~Z., Lin, Z., Padhy, S., Tran, D., Bedrax{-}Weiss, T., and
  Lakshminarayanan, B.
\newblock Simple and principled uncertainty estimation with deterministic deep
  learning via distance awareness.
\newblock \emph{CoRR}, abs/2006.10108, 2020.

\bibitem[Maddox et~al.(2019)Maddox, Izmailov, Garipov, Vetrov, and
  Wilson]{swag}
Maddox, W.~J., Izmailov, P., Garipov, T., Vetrov, D.~P., and Wilson, A.~G.
\newblock A simple baseline for bayesian uncertainty in deep learning.
\newblock In \emph{Advances in Neural Information Processing Systems},
  volume~32. Curran Associates, Inc., 2019.

\bibitem[Malinin \& Gales(2018)Malinin and Gales]{single-pass-ood}
Malinin, A. and Gales, M.
\newblock Predictive uncertainty estimation via prior networks, 2018.

\bibitem[Nado et~al.(2021)Nado, Band, Collier, Djolonga, Dusenberry, Farquhar,
  Filos, Havasi, Jenatton, Jerfel, Liu, Mariet, Nixon, Padhy, Ren, Rudner, Wen,
  Wenzel, Murphy, Sculley, Lakshminarayanan, Snoek, Gal, and
  Tran]{uncertainty-baselines}
Nado, Z., Band, N., Collier, M., Djolonga, J., Dusenberry, M., Farquhar, S.,
  Filos, A., Havasi, M., Jenatton, R., Jerfel, G., Liu, J., Mariet, Z., Nixon,
  J., Padhy, S., Ren, J., Rudner, T., Wen, Y., Wenzel, F., Murphy, K., Sculley,
  D., Lakshminarayanan, B., Snoek, J., Gal, Y., and Tran, D.
\newblock {Uncertainty Baselines}: Benchmarks for uncertainty \& robustness in
  deep learning.
\newblock \emph{arXiv preprint arXiv:2106.04015}, 2021.

\bibitem[Papernot et~al.(2016{\natexlab{a}})Papernot, McDaniel, and
  Goodfellow]{bboxattack2}
Papernot, N., McDaniel, P., and Goodfellow, I.
\newblock Transferability in machine learning: from phenomena to black-box
  attacks using adversarial samples, 2016{\natexlab{a}}.

\bibitem[Papernot et~al.(2016{\natexlab{b}})Papernot, McDaniel, Goodfellow,
  Jha, Celik, and Swami]{bboxattack}
Papernot, N., McDaniel, P., Goodfellow, I., Jha, S., Celik, Z.~B., and Swami,
  A.
\newblock Practical black-box attacks against machine learning,
  2016{\natexlab{b}}.

\bibitem[Pope et~al.(2021)Pope, Zhu, Abdelkader, Goldblum, and
  Goldstein]{imagemanifold2}
Pope, P., Zhu, C., Abdelkader, A., Goldblum, M., and Goldstein, T.
\newblock The intrinsic dimension of images and its impact on learning, 2021.

\bibitem[Ritter et~al.(2018)Ritter, Botev, and Barber]{laplace_oodadv}
Ritter, H., Botev, A., and Barber, D.
\newblock A scalable laplace approximation for neural networks.
\newblock In \emph{International Conference on Learning Representations}, 2018.

\bibitem[Ruderman(1994)]{imagemanifold1}
Ruderman, D.~L.
\newblock The statistics of natural images.
\newblock \emph{Network: Computation in Neural Systems}, 5\penalty0
  (4):\penalty0 517--548, 1994.
\newblock \doi{10.1088/0954-898X_5_4_006}.

\bibitem[Sheikholeslami et~al.(2019)Sheikholeslami, Jain, and
  Giannakis]{thresholding_mi_adv}
Sheikholeslami, F., Jain, S., and Giannakis, G.~B.
\newblock Minimum uncertainty based detection of adversaries in deep neural
  networks, 2019.

\bibitem[Smith \& Gal(2018)Smith and Gal]{understanding_measures}
Smith, L. and Gal, Y.
\newblock Understanding measures of uncertainty for adversarial example
  detection, 2018.

\bibitem[Strauss et~al.(2017)Strauss, Hanselmann, Junginger, and
  Ulmer]{ensemblesrobust}
Strauss, T., Hanselmann, M., Junginger, A., and Ulmer, H.
\newblock Ensemble methods as a defense to adversarial perturbations against
  deep neural networks, 2017.

\bibitem[van Amersfoort et~al.(2020)van Amersfoort, Smith, Teh, and Gal]{duq}
van Amersfoort, J., Smith, L., Teh, Y.~W., and Gal, Y.
\newblock Uncertainty estimation using a single deep deterministic neural
  network, 2020.

\bibitem[Wenzel et~al.(2020)Wenzel, Snoek, Tran, and Jenatton]{hyperparamde}
Wenzel, F., Snoek, J., Tran, D., and Jenatton, R.
\newblock Hyperparameter ensembles for robustness and uncertainty
  quantification, 2020.

\bibitem[Xia et~al.(2021)Xia, Han, and Mascolo]{unccomp}
Xia, T., Han, J., and Mascolo, C.
\newblock Benchmarking uncertainty qualification on biosignal classification
  tasks under dataset shift.
\newblock \emph{CoRR}, abs/2112.09196, 2021.

\end{thebibliography}
\bibliographystyle{icml2022}

\appendix
\section{Model Architectures} 
All models share basic architecture. For half moons and toy manifold experiments, all models except MC Dropout use a feed-forward neural network with two hidden layers with 20 nodes per layer and ReLU activation. MC Dropout uses 128 nodes per layer. For MNIST we use a simple convolutional network with three layers with 64, 128, and 128 filters respectively and a 3$\times$3 kernel. Each layer is passed through batch-normalization layer followed by ReLU activation and max-pooling with a 2$\times$2 kernel. SNGP employs an additive residual connection after the  batch-normalization layers. MC Dropout uses filter-wise dropout before each pooling layer.

\section{Hyperparameter Tuning}
Where possible, models and the ideal hyperparameters were taking directly from the referenced papers. This was true for deep ensembles and DUQ taken from the DUQ paper \cite{duq}. Otherwise, models were tuned using a grid search over model-specific parameters, following the tuning procedures specific to that model if they exist, e.g. the spec-norm value in SNGP. The halfmoons case is straight-forward to achieve good accuracy on and requires only standard parameters for each model. Below we briefly summarize the final model parameters in the MNIST and 1D Toy Manifold cases.

\subsection{MNIST}

\textbf{BCNN} RMSprop optimizer with learning rate of 0.0001 and batch size of 64. We compute the distribution for each parameter in this model with kernel divergence set to a KL divergence function. 

\textbf{DE} SGD optimizer with momentum of 0.9, weight decay of 0.00001, learning rate 0.00001, and batch size 32. 30 models are trained at 5 epochs each.

\textbf{MC Dropout} Adam optimizer, batch size of 256, learning rate of 0.0001, filter-wise dropout rate of 10\%. \textit{T} = 100 forward passes were then averaged to get the final softmaxed probability vector.

\textbf{SWAG} SGD optimizer with momentum of 0.9, weight decay of 0.00001, learning rate 0.00001, and batch size 32. Model exploration proceeds for 30 epochs at constant learning rate 0.1 and low-rank matrix includes the last 10 epochs. 30 samples are taken from the parameterized Gaussian.

\textbf{DUQ} SGD optimizer with learning rate 0.0001, momentum of 0.9, and weight decay 0.00001. 

\textbf{SNGP} SGD with a momentum of 0.9 and learning rate of 0.01, batch size of 64, spec\_norm\_iteration of 1, spec\_norm\_bound of 1.

\subsection{1D Toy Manifold}

\textbf{BNN} For the BNN we use HMC with 1 chain, 1000 warm-up steps while sampling 50 models each time. The sampler is tuned to start with an adaptable step size of $10^{-6}$ and the acceptance probability is set as 0.95. The max tree depth of the sampler is set to 5.

\textbf{DE} SGD optimizer with learning rate 0.001. 20 models are trained at 500 epochs each.

\textbf{MC Dropout} Adam optimizer with a learning rate of 0.0001, clipnorm of 0.5, batch size of 32, dropout rate of 20\%, 650 epochs regularized with early stopping on the loss with a patience of 50.

\begin{figure}
    \centering
    \includegraphics[width=0.4\columnwidth]{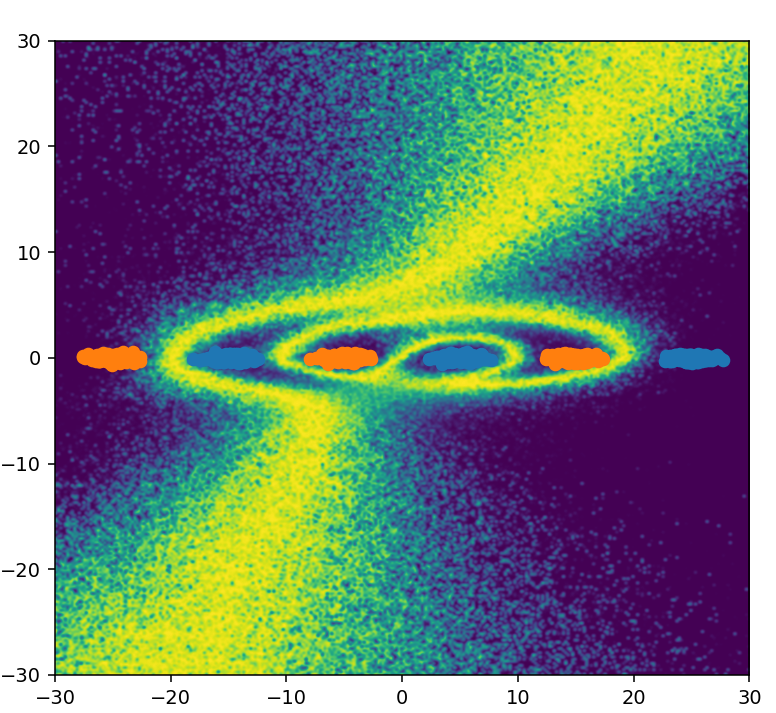}
    \includegraphics[width=0.45\columnwidth]{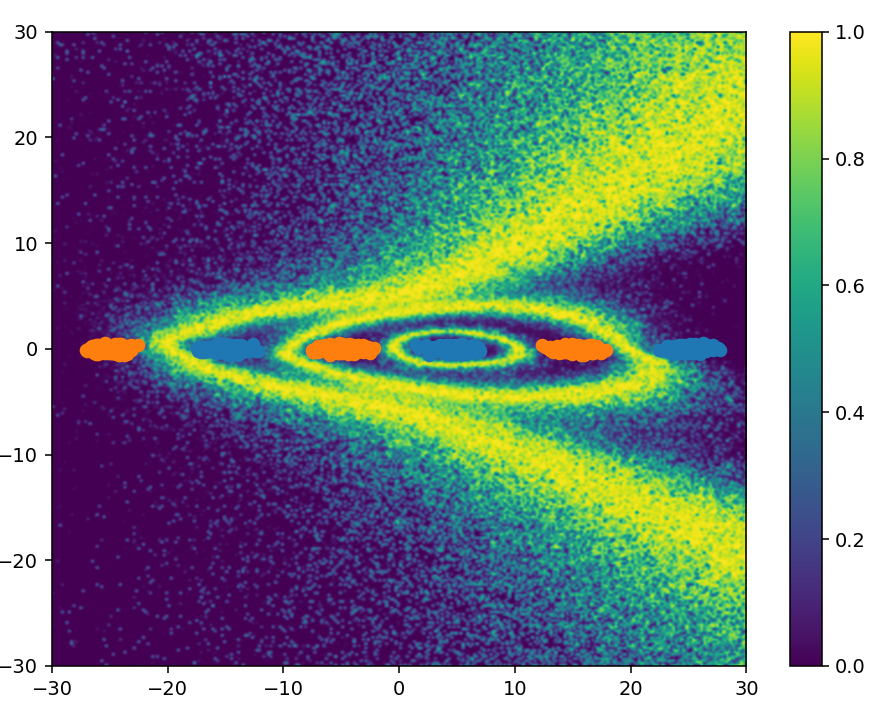}
    \caption{Epistemic Uncertainties for two different random restarts of MC Dropout on the Toy Manifold Data. All procedures are held the same except for the tensorflow random seed, which is set to 12345 on the left and 99999 on the right.}
    \label{mc-random-restarts}
\end{figure}

The uncertainties and decision boundaries produced for MC Dropout showed to be susceptible to random restarts. Figure~\ref{mc-random-restarts} demonstrates the decision boundaries and epistemic uncertainties for two MC Dropout models trained on the toy manifold data using the above procedure, only differing by setting different random seeds. 

\textbf{SWAG} SGD optimizer with learning rate 0.001. Model is pre-trained for 600 epochs with exploration for 30 epochs at learning rate 0.033 and low-rank matrix recording the last 5 epochs.

\textbf{DUQ} Adam optimizer with learning rate 0.0001, batch size 32, and trained for 500 epochs.

\textbf{SNGP} Adam with a learning rate of 0.0001, batch size of 32, spec\_norm\_iteration of 1, spec\_norm\_bound of 0.9, dropout rate of 10\%, 750 epochs regularized with early stopping on the loss with a patience of 50.

\section{Uncertainty Metrics}
\label{umformulae}
Let $M$ be the number of component models in an ensemble and $C$ be the number of classes.

The \textit{aleatoric entropy} is computed as:
$$
\text{H}_a(x_n) = -\frac{1}{M}\sum_m^M\sum_c^Cp(y_c|x_n,W_m)\log(p(y_c|x_n,W_m))
$$

The \textit{epistemic entropy} is computed as:
$$
\text{H}_e(x_n) = -\sum_{c}^C\bar{p}_c\log(\bar{p}_c)
$$
where,
$$
\bar{p}_c = \frac{1}{M}\sum_m^Mp(y_c|x_n,W_m)
$$

The \textit{epistemic KL uncertainty} is computed as:
$$
\text{KL}_e(x_n) = \frac{1}{P}\sum_{m',m''}^{P} KL\left(p(y|x_n,W_{m'}), p(y|x_n,W_{m''})\right) 
$$
where $P$ is the $\frac{M!}{(M-2)!}$ combinations of models. We do not restrict to unique combinations due to the asymmetry of KL-divergence. 

For aleatoric entropy in DUQ we have $M=1$ and
$$p(y|x_n,W) = \text{softmax}(\log(f_W(x_n))))$$

\section{Adversarial Robustness}
\label{adv-appendix}
To generate adversarial examples in both the halfmoons and MNIST cases, we design a proxy model ($f_p$) trained to approximate the target model's ($f_t$) decision boundary. Specifically, we have a trained target model that produces predictions $\hat{y}_t$ that are optimized by minimizing the loss with respect to the true labels, $min\;\mathcal L(\hat{y}_t, y_{true})$. The target model learns by matching the target model's predictions, i.e. $min\;\mathcal L(\hat{y}_p, \hat{y}_t)$. With the trained proxy we then perform an iterative, L2 projected gradient descent (L2-PGD) attack.\footnote{Implementation adapted from \href{https://adversarial-ml-tutorial.org/adversarial_examples/}{https://adversarial-ml-tutorial.org/adversarial\_examples}} Specifically, we learn a perturbation $\delta$ over a number of iterations where in each iteration we update $\delta$ such that,
$$
\delta_{t+1} = \mathcal P(\delta_{t} + \alpha\nabla_{\delta}\mathcal L(f_p(x + \delta_t), \hat{y}_t))
$$

where $\alpha$ is the learning rate, $\mathcal P$ is the projection onto the L2-ball, and $\delta$ is bounded by a maximum perturbation $\epsilon$.

Iterative PGD attacks have three hyperparameters: maximum allowable perturbation ($\epsilon$), step size ($\alpha$), and number of gradient steps. In order to mimic real-world application of a proxy model based attack, we tuned the parameters in order to keep the degree to which the proxy model was fooled consistent. As a final check, we track the average L2 norm of the perturbations. For measuring adversarial robustness using SNGP we utilize the mean-field approximation to the posterior predictive \cite{sngp, uncertainty-baselines}, even though we take samples for uncertainty approximation. Empirically the difference in accuracy was not significant.

To generate on/off-manifold adversaries for the toy manifold data we add a perturbation, $\delta$, to the samples where $\delta \sim \mathcal{N}(\epsilon, 0.5)$. $\epsilon$ serves as a strength parameter for our "attack." Additionally we randomly select with probability 0.5, for each data point, whether the perturbation will be applied in the positive or negative direction. To create on-manifold adversaries we add the perturbation to the x-coordinate, and to create off-manifold adversaries we add the perturbation to the y-coordinate. We show accuracy as a function of $\epsilon$ for each model for on/off-manifold adversaries in \hyperref[toy-mani-robustness]{Figure 6}.

\begin{table}[h!]
\setlength{\tabcolsep}{2pt}
\caption{Adversarial Robustness and Hyperparameters (half moons)}
\label{adv-table}
\vskip 0.15in
\begin{center}
\begin{small}
\begin{sc}
\begin{tabular}{lcccccc}
\toprule
Model & Proxy Acc. & Adv. Acc. &$\epsilon$ & $\alpha$ & Iter. \\
\midrule
BNN  & 32.14\% & 39.87\% & 0.6 & 0.01 & 40 &\\
DE & 39.67\%&39.58\%& 0.6& 0.01&40\\
MC Drop.  & 32.80\% & 36.90\% & 0.6 & 0.008 & 40  \\
SWAG    & 35.67\%&39.58\%& 0.6& 0.01&40        \\
DUQ      & 38.33\%&30.56\%& 0.6& 0.01&40 \\
SNGP      & 30.09\% & 36.20\% & 0.6 & 0.01  & 40 \\
\bottomrule
\end{tabular}
\end{sc}
\end{small}
\end{center}
\vskip -0.1in
\end{table}

\begin{table}[h!]
\setlength{\tabcolsep}{2pt}
\caption{Adversarial Robustness and Hyperparameters (MNIST)}
\label{mnist-adv-table}
\vskip 0.15in
\begin{center}
\begin{small}
\begin{sc}
\begin{tabular}{lcccccc}
\toprule
Model & Proxy Acc. & Adv. Acc. &$\epsilon$ & $\alpha$ & Iter.&L2 \\
\midrule
BNN & 5.89\%  & 90.20\% & 25 & 0.10 & 50 & 2.70 \\
DE & 3.50\% & 85.92\% & 25& 0.10 & 50 & 10.92\\
MC Drop. & 6.77\% & 90.39\% & 25 & 0.06 & 50 & 3.30 \\
SWAG & 8.47\%& 37.76\% & 25 & 0.12 & 50 & 12.41 \\
DUQ & 2.16\%& 43.53\% & 25 & 0.10 & 50 & 8.74 \\
SNGP & 3.45\% & 85.48\% & 25 & 0.07 & 50 & 4.18 \\
\bottomrule
\end{tabular}
\end{sc}
\end{small}
\end{center}
\vskip -0.1in
\end{table}

\begin{figure}[h!]
    \label{halfmoons-adv-proxy-illustration}
     \centering
     \begin{subfigure}[b]{0.48\columnwidth}
         \centering
         \includegraphics[width=\columnwidth]{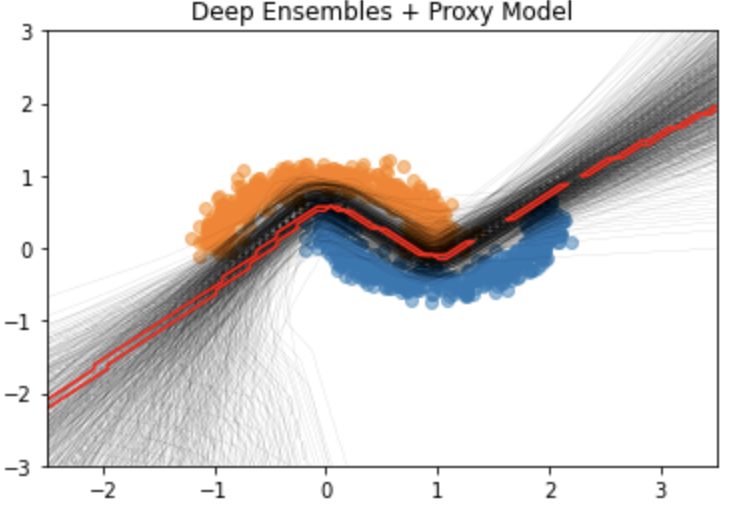}
     \end{subfigure}
     \begin{subfigure}[b]{0.48\columnwidth}
         \centering
         \includegraphics[width=\columnwidth]{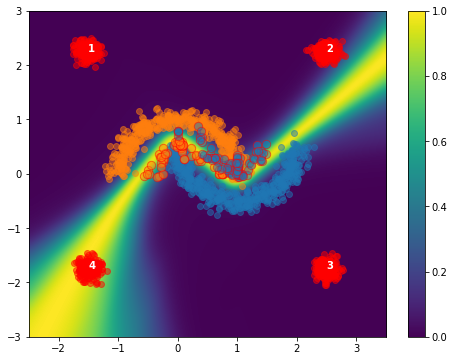}
     \end{subfigure}
     \begin{subfigure}[b]{\columnwidth}
         \centering
         \includegraphics[width=\columnwidth]{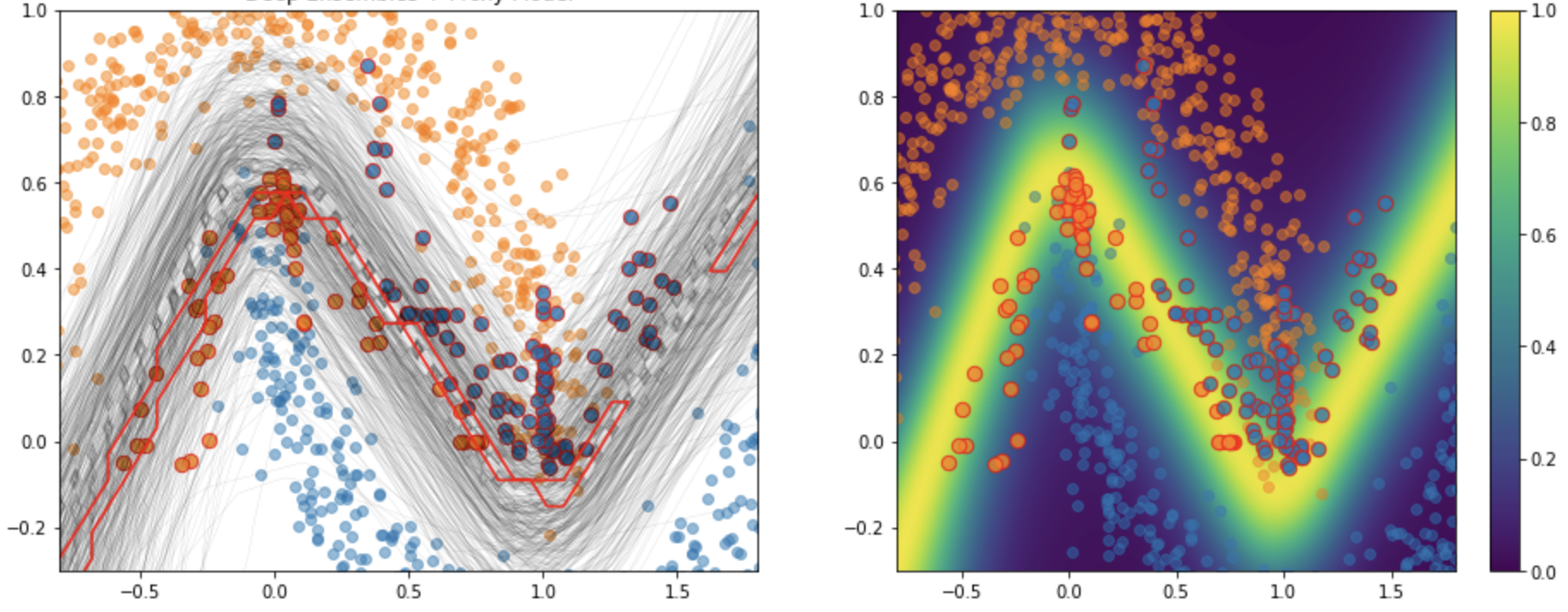}
     \end{subfigure}
     \caption{Illustration of a black-box attack on deep ensembles using a standard Neural Network as a proxy model. Top left: The many decision boundaries in the deep ensemble (black) with the proxy model's approximating boundary (red). Top right: Deep Ensemble epistemic uncertainty with the generated adversaries plotted. Brightly colored points indicate that this adversary belonging to its color was incorrectly classified as the other.}
\end{figure}

\begin{figure}[h!]
    \label{toy-mani-robustness}
     \centering
     \includegraphics[width=\columnwidth]{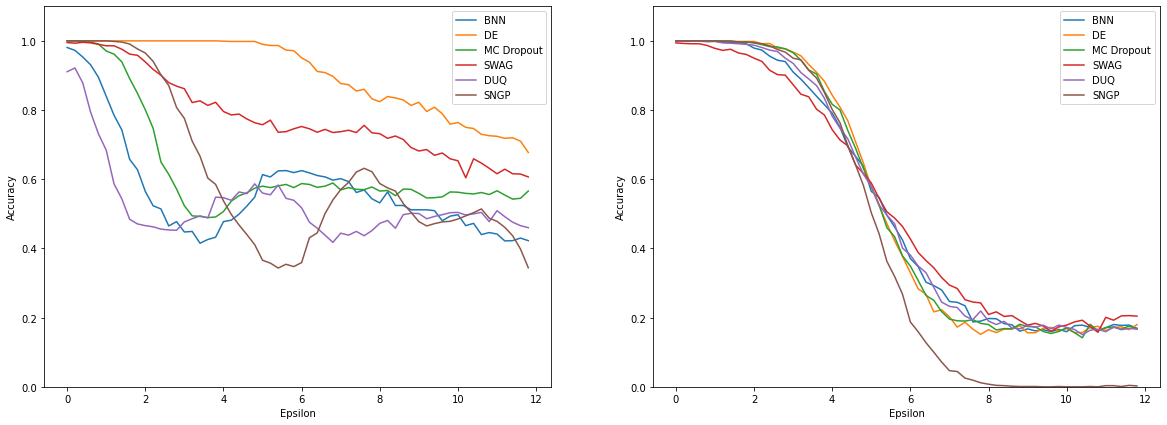}
     
     \caption{Adversarial accuracy (robustness) in the toy manifold setting. Adversaries are generated by perturbing data on- or off-manifold for increasing values of epsilon. Off-manifold perturbations (left) elicit discrepancies in robustness similar to MNIST, though results vary for models that fluctuate by random restart. All models show consistent behavior for on-manifold perturbations (right).}
\end{figure}

In \hyperref[halfmoons-adv-proxy-illustration]{Figure 5} we show a visualization of a proxy model and generated adversaries on a deep ensemble. Similar visualizations can be generated for the other models, but we omit them for brevity.

\section{Halfmoons Uncertainties}
\label{appendix-halfmoons-uncertainties}

We show here the aleatoric, epistemic, and KL uncertainties for each model. DUQ has no notion of KL uncertainty since its epistemic uncertainty is defined explicitly without an ensemble of predictions. In SNGP we use the mean-field approximation for producing the softmax probabilities that feed into the aleatoric uncertainty, but we use samples from the GP posterior to create an ensemble of predictions to calculate epistemic and KL uncertainties. We take 10 samples as was specified in the original paper \cite{sngp}.

\begin{figure}[h!]
     \centering
     \begin{subfigure}[b]{0.3\columnwidth}
         \centering
         \includegraphics[width=\columnwidth]{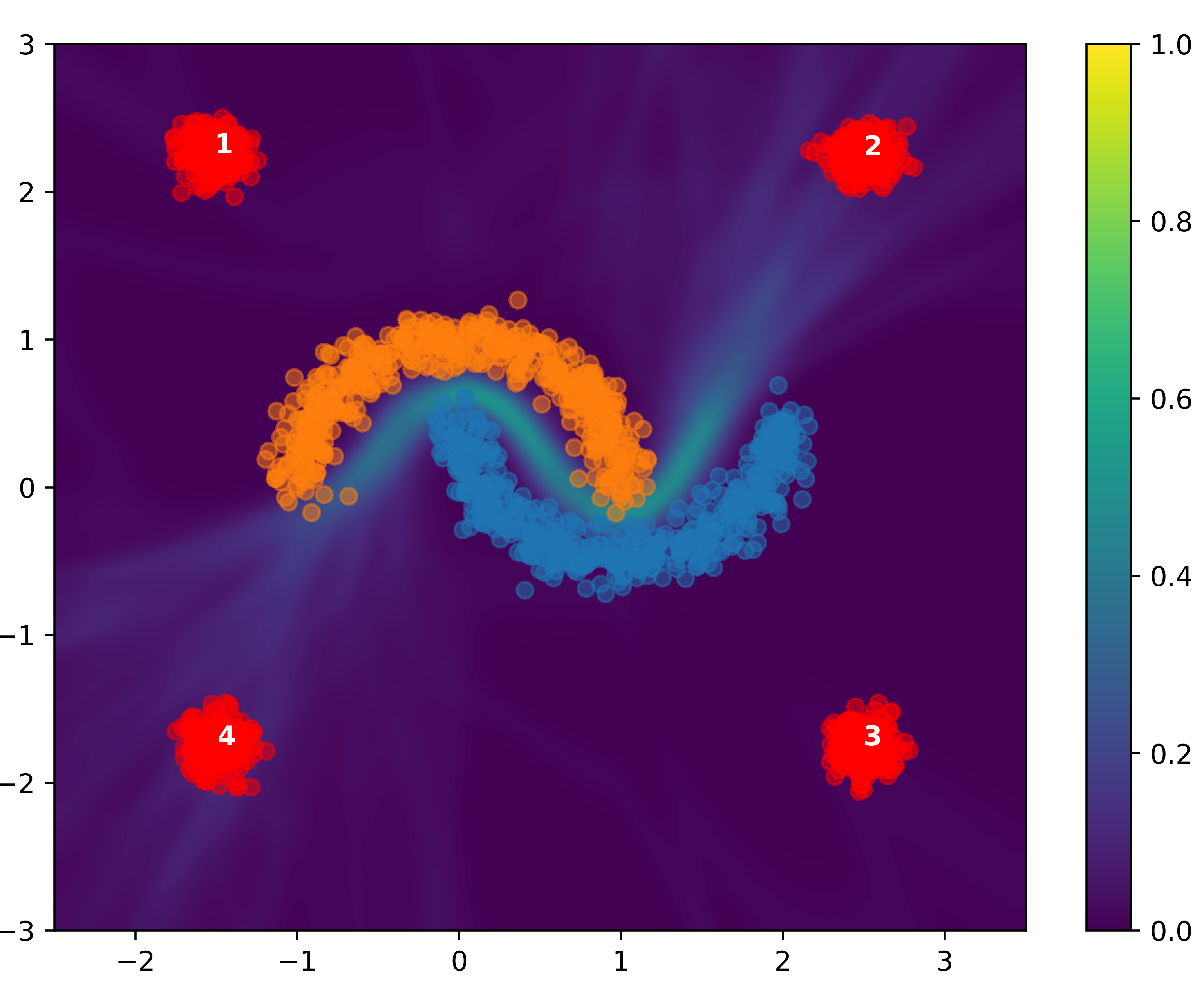}
         \caption{BNN Aleatoric}
     \end{subfigure}
     \hfill
     \begin{subfigure}[b]{0.3\columnwidth}
         \centering
         \includegraphics[width=\columnwidth]{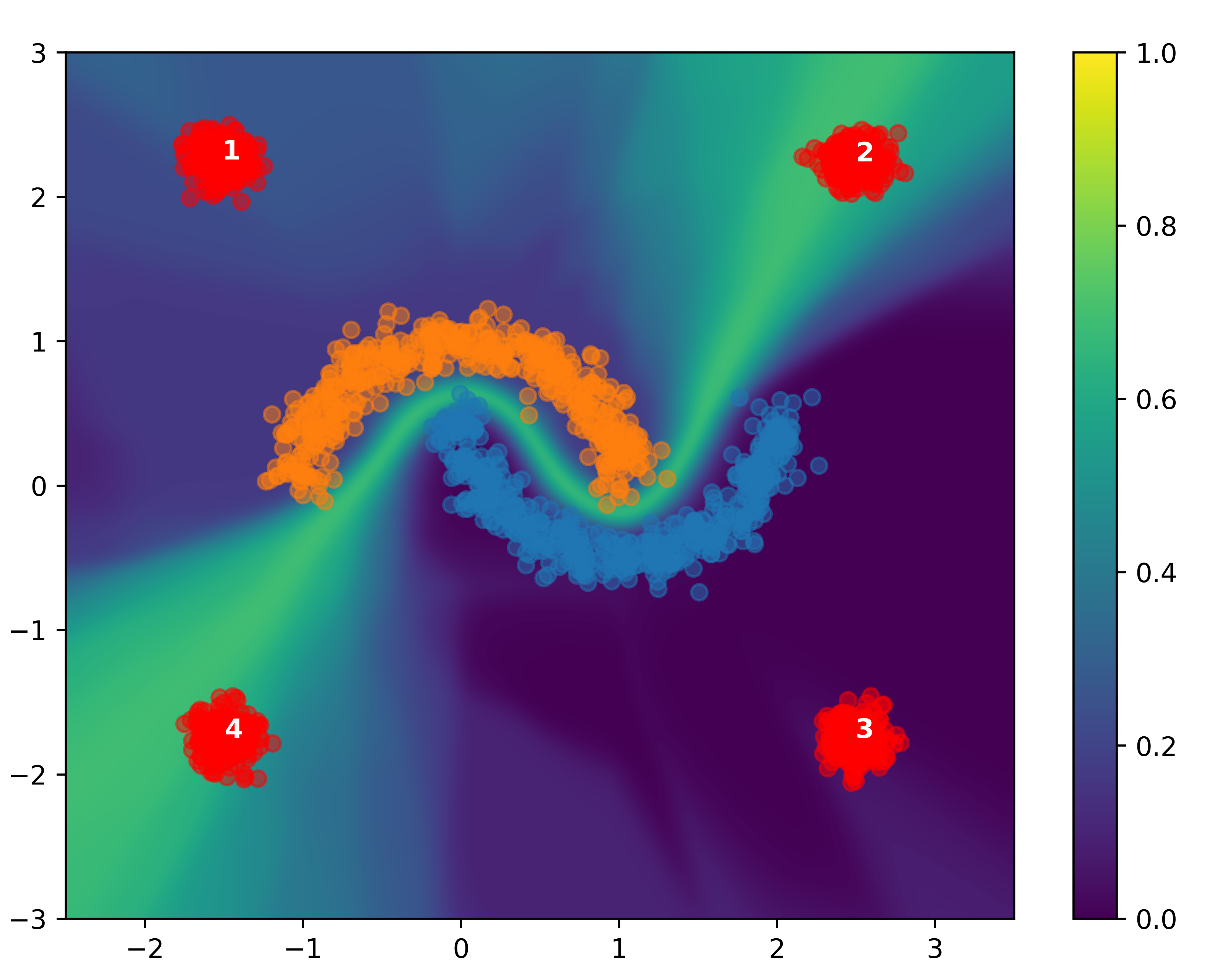}
         \caption{BNN Epistemic}
     \end{subfigure}
     \hfill
     \begin{subfigure}[b]{0.3\columnwidth}
         \centering
         \includegraphics[width=\columnwidth]{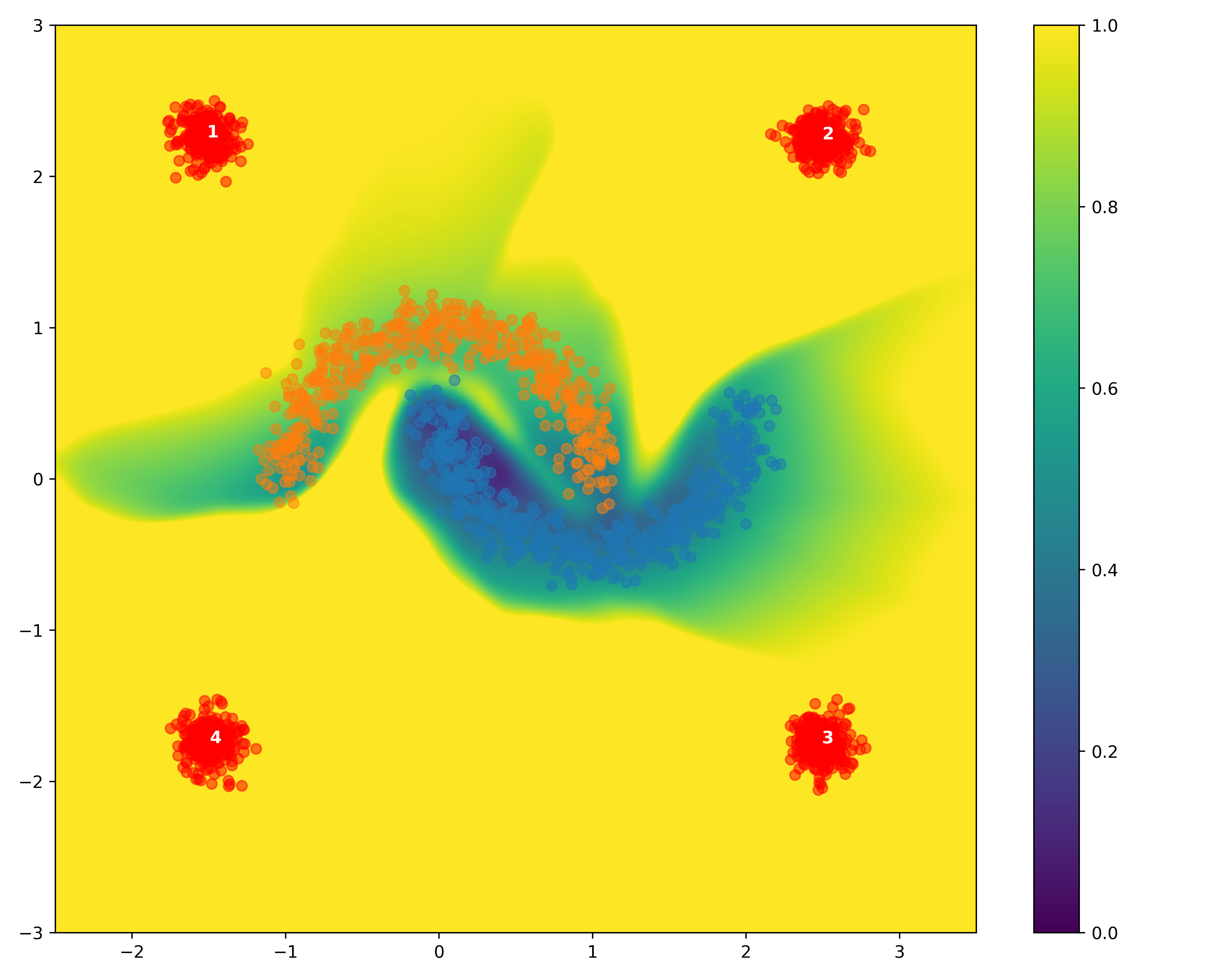}
         \caption{BNN KL}
     \end{subfigure}
\end{figure}

\begin{figure}[h!]
     \begin{subfigure}[b]{0.3\columnwidth}
         \centering
         \includegraphics[width=\columnwidth]{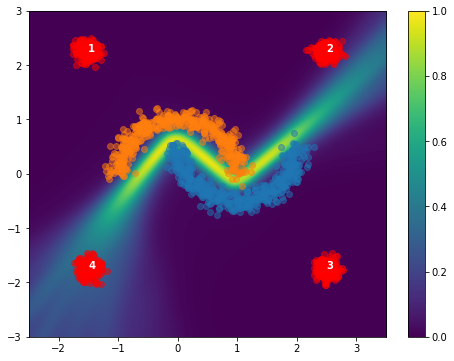}
         \caption{DE Aleatoric}
     \end{subfigure}
    \hfill
    \begin{subfigure}[b]{0.3\columnwidth}
         \centering
         \includegraphics[width=\columnwidth]{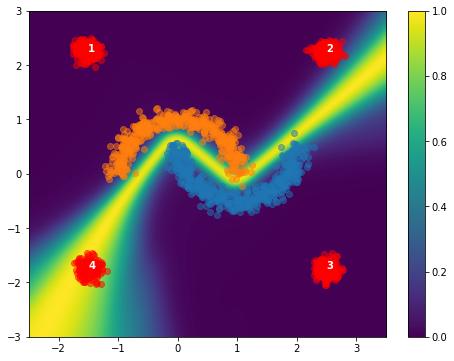}
         \caption{DE Epistemic}
     \end{subfigure}
     \hfill
     \begin{subfigure}[b]{0.3\columnwidth}
         \centering
         \includegraphics[width=\columnwidth]{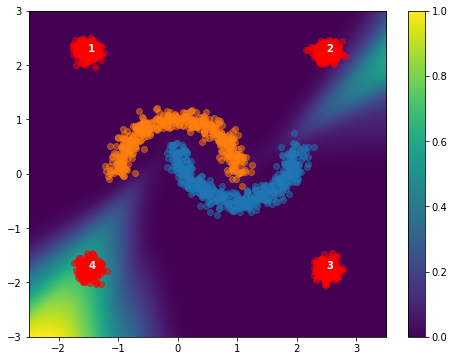}
         \caption{DE KL}
     \end{subfigure}
     \hfill
\end{figure}

\begin{figure}[h!]
     \begin{subfigure}[b]{0.3\columnwidth}
         \centering
         \includegraphics[width=\columnwidth]{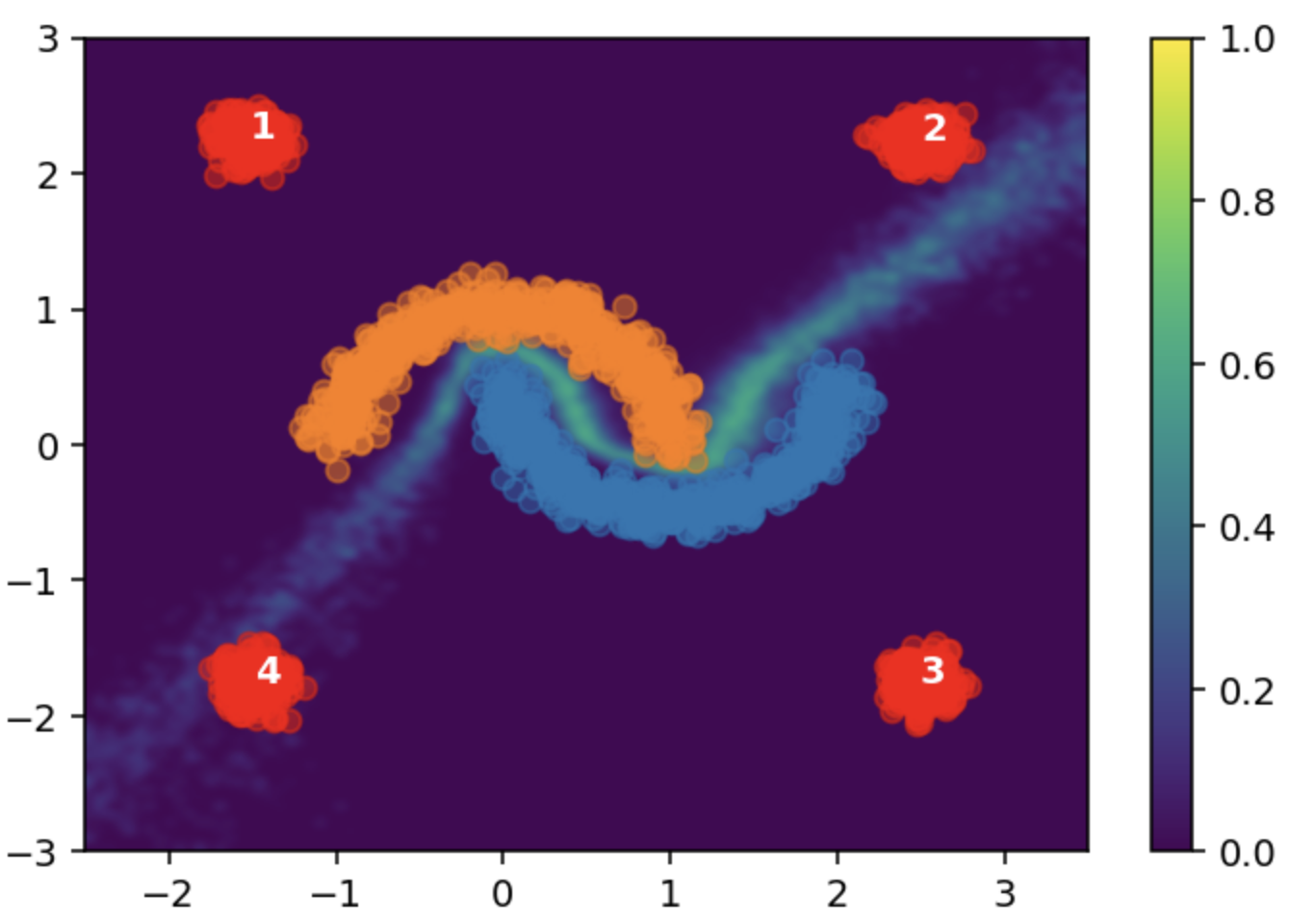}
         \caption{\tiny{MC Dropout Aleatoric}}
     \end{subfigure}
    \hfill
    \begin{subfigure}[b]{0.3\columnwidth}
         \centering
         \includegraphics[width=\columnwidth]{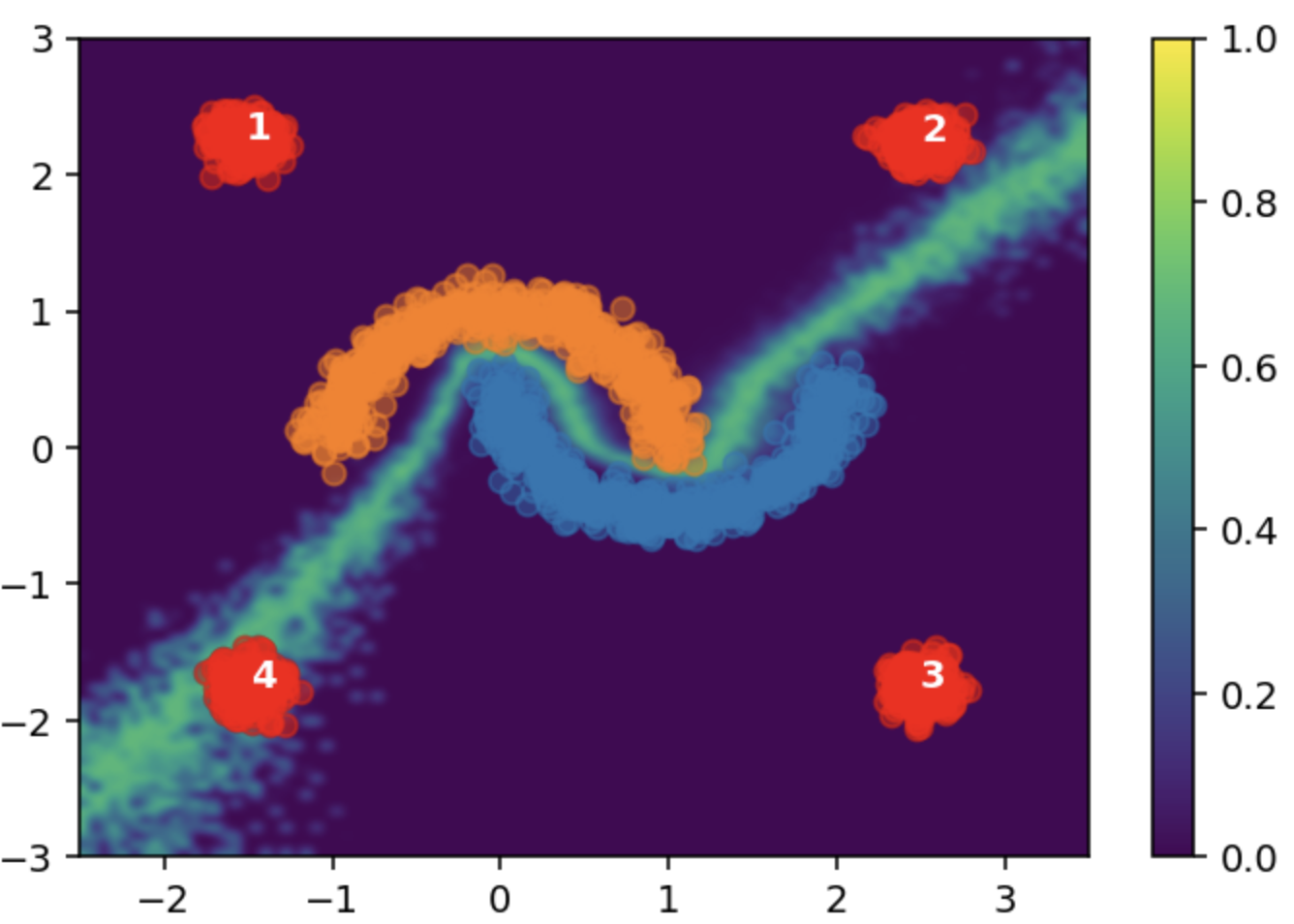}
         \caption{\tiny{MC Dropout Epistemic}}
     \end{subfigure}
     \hfill
     \begin{subfigure}[b]{0.3\columnwidth}
         \centering
         \includegraphics[width=\columnwidth]{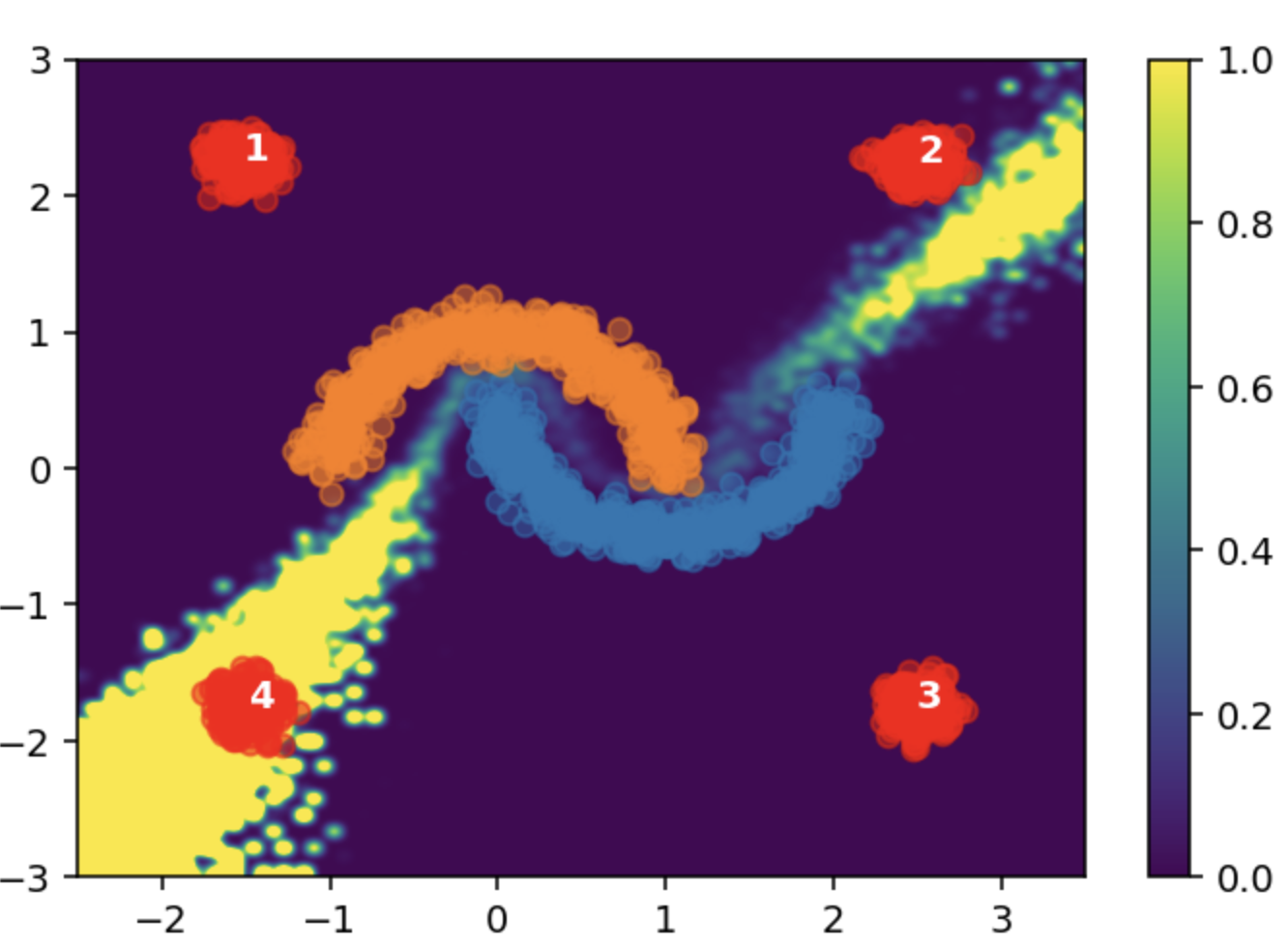}
         \caption{MC Dropout KL}
     \end{subfigure}
     \hfill
\end{figure}

\begin{figure}[h!]
     \begin{subfigure}[b]{0.3\columnwidth}
         \centering
         \includegraphics[width=\columnwidth]{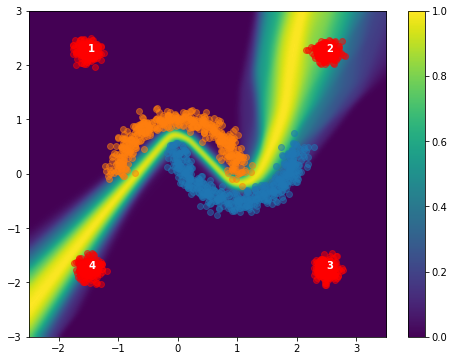}
         \caption{\tiny{SWAG Aleatoric}}
     \end{subfigure}
    \hfill
    \begin{subfigure}[b]{0.3\columnwidth}
         \centering
         \includegraphics[width=\columnwidth]{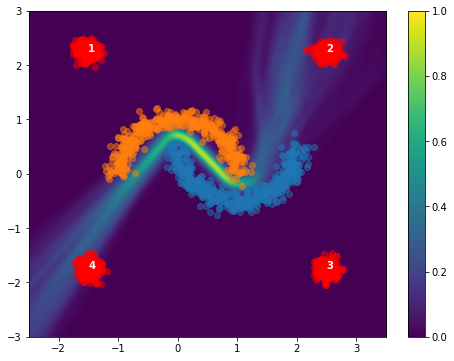}
         \caption{\tiny{SWAG Epistemic}}
     \end{subfigure}
     \hfill
     \begin{subfigure}[b]{0.3\columnwidth}
         \centering
         \includegraphics[width=\columnwidth]{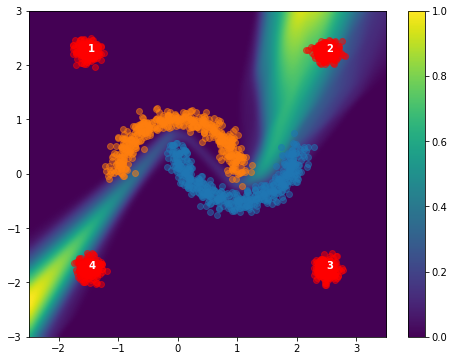}
         \caption{SWAG KL}
     \end{subfigure}
     \hfill
\end{figure}

\begin{figure}[h!]
     \begin{subfigure}[b]{0.3\columnwidth}
         \centering
         \includegraphics[width=\columnwidth]{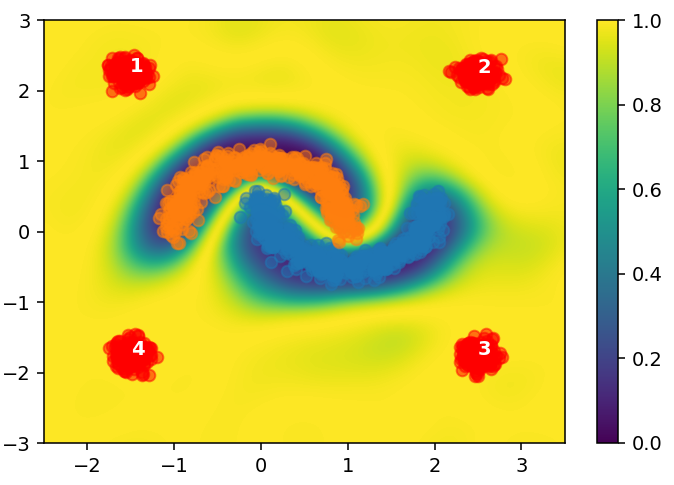}
         \caption{\tiny{SNGP Aleatoric}}
     \end{subfigure}
    \hfill
    \begin{subfigure}[b]{0.3\columnwidth}
         \centering
         \includegraphics[width=\columnwidth]{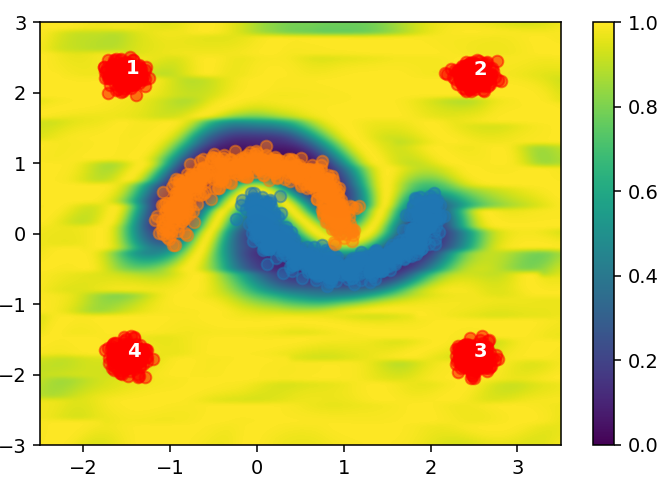}
         \caption{\tiny{SNGP Epistemic}}
     \end{subfigure}
     \hfill
     \begin{subfigure}[b]{0.3\columnwidth}
         \centering
         \includegraphics[width=\columnwidth]{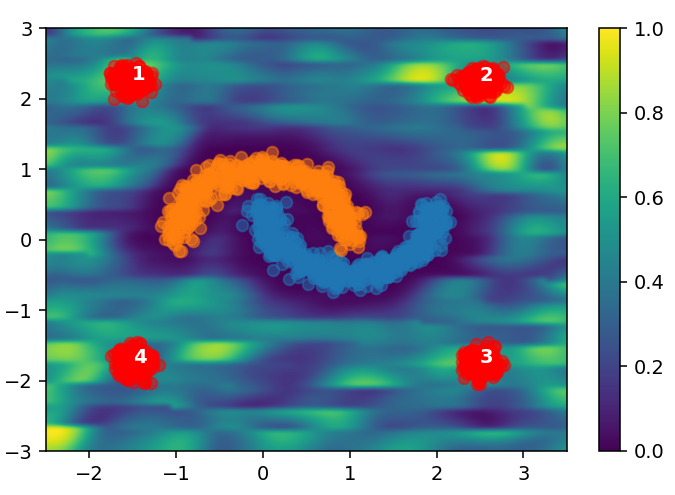}
         \caption{\tiny{SNGP KL}}
     \end{subfigure}
     \hfill
\end{figure}

\begin{figure}[h!]
     \begin{subfigure}[b]{0.4\columnwidth}
         \centering
         \includegraphics[width=\columnwidth]{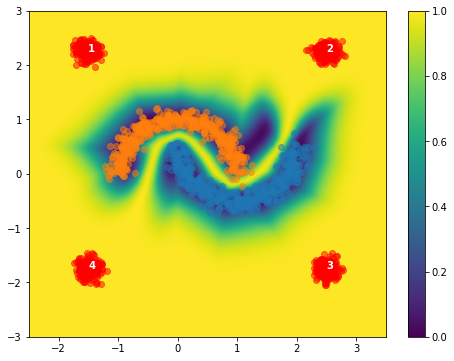}
         \caption{DUQ Aleatoric}
     \end{subfigure}
    \hfill
    \begin{subfigure}[b]{0.4\columnwidth}
         \centering
         \includegraphics[width=\columnwidth]{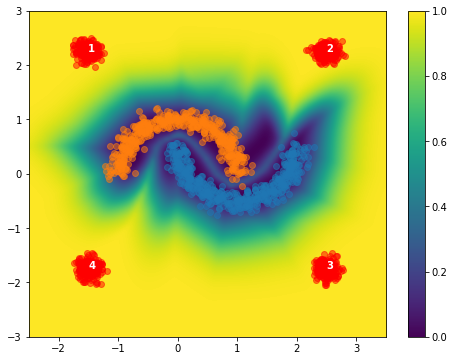}
         \caption{DUQ Epistemic}
     \end{subfigure}
\end{figure}
\section{Toy Manifold Uncertainties}
\label{appendix-toy-manifold-uncertainties}

We show here the aleatoric, epistemic, and KL uncertainties for each model. DUQ has no notion of KL uncertainty since its epistemic uncertainty is defined explicitly without an ensemble of predictions. In SNGP we use the mean-field approximation for producing the softmax probabilities that feed into the aleatoric uncertainty, but we use samples from the GP posterior to create an ensemble of predictions to calculate epistemic and KL uncertainties. We take 10 samples as was specified in the original paper \cite{sngp}.

\begin{figure}[h!]
     \centering
     \begin{subfigure}[b]{0.3\columnwidth}
         \centering
         \includegraphics[width=\columnwidth]{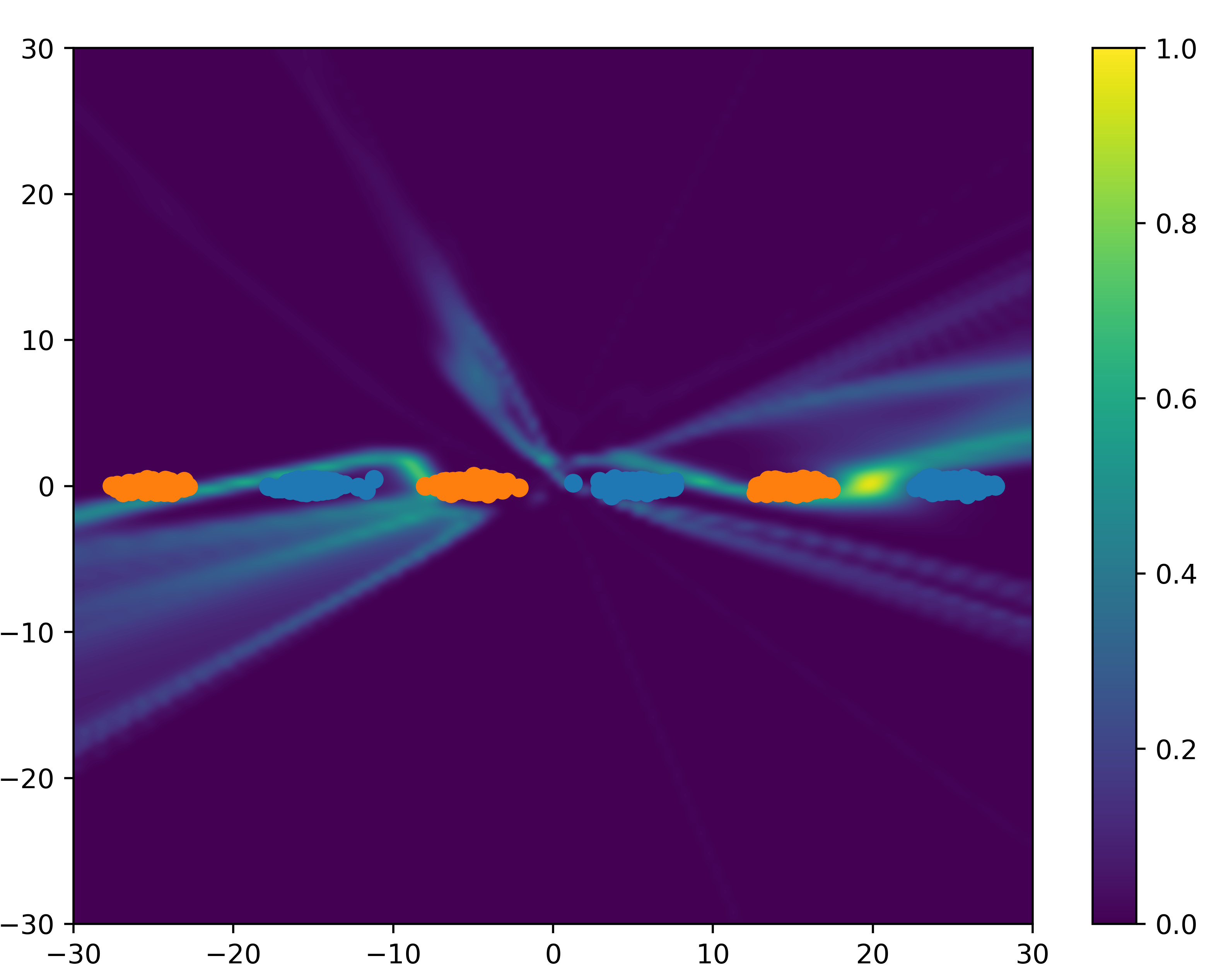}
         \caption{BNN Aleatoric}
     \end{subfigure}
     \hfill
     \begin{subfigure}[b]{0.3\columnwidth}
         \centering
         \includegraphics[width=\columnwidth]{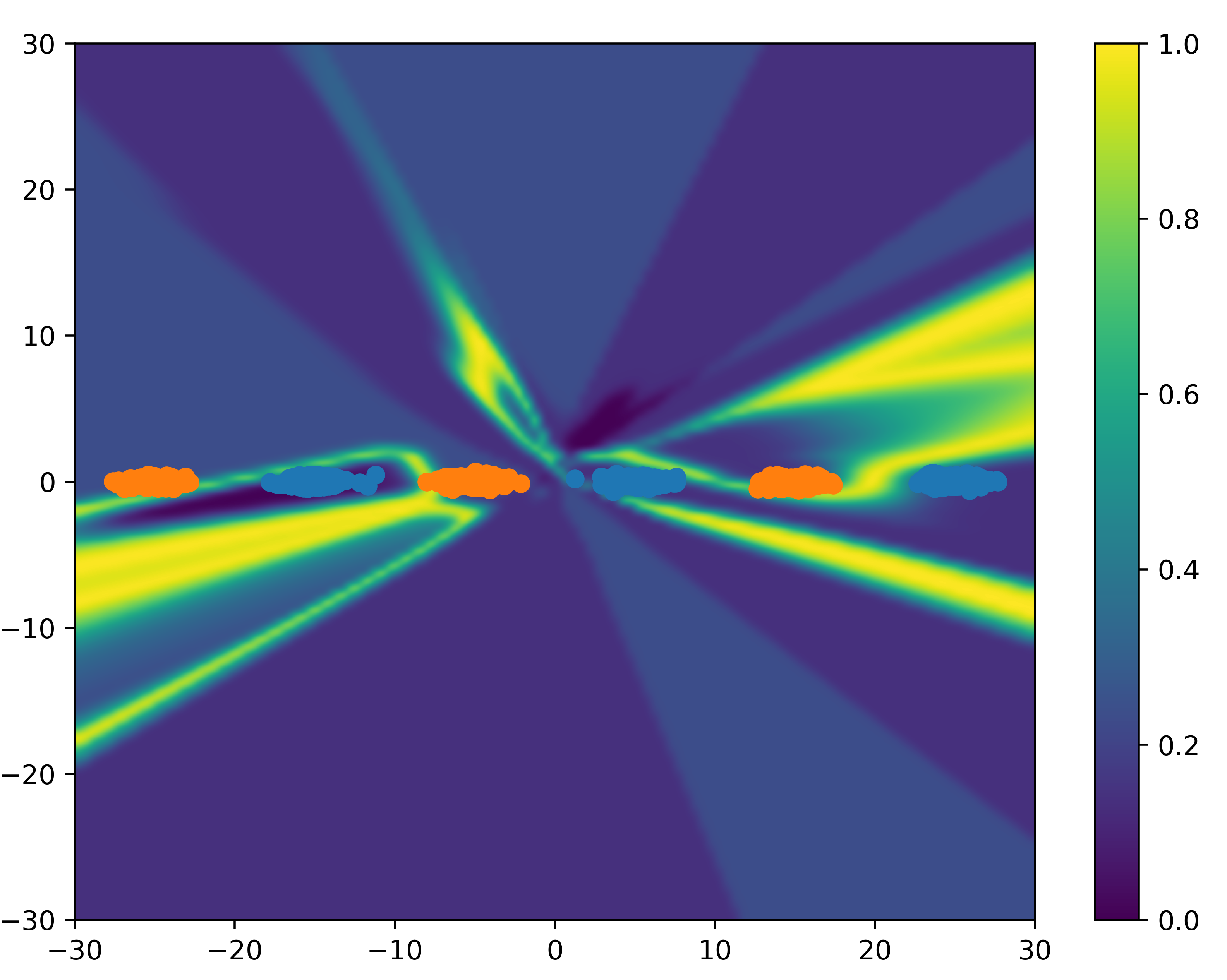}
         \caption{BNN Epistemic}
     \end{subfigure}
     \hfill
     \begin{subfigure}[b]{0.3\columnwidth}
         \centering
         \includegraphics[width=\columnwidth]{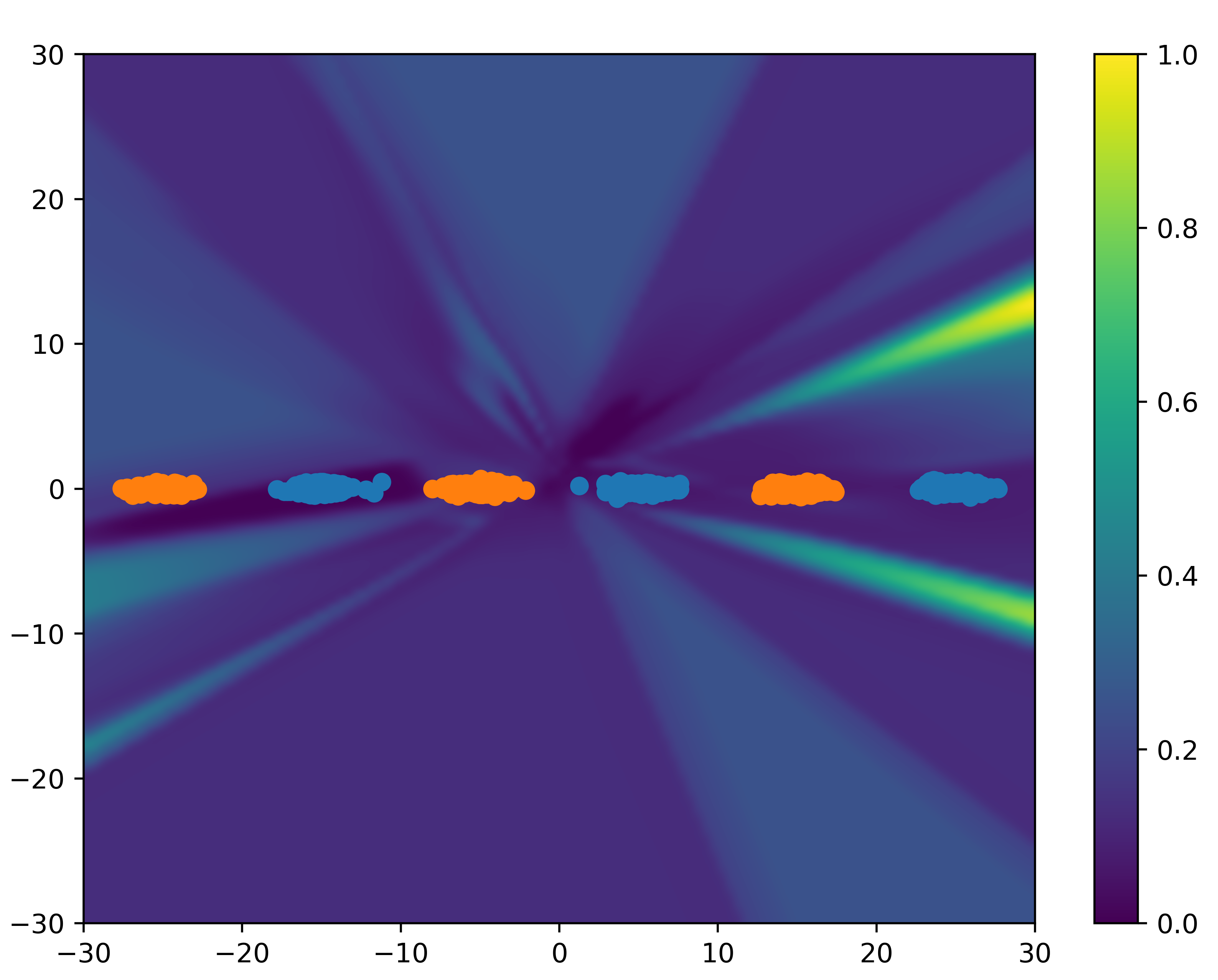}
         \caption{BNN KL}
     \end{subfigure}
\end{figure}

\begin{figure}[h!]
     \begin{subfigure}[b]{0.3\columnwidth}
         \centering
         \includegraphics[width=\columnwidth]{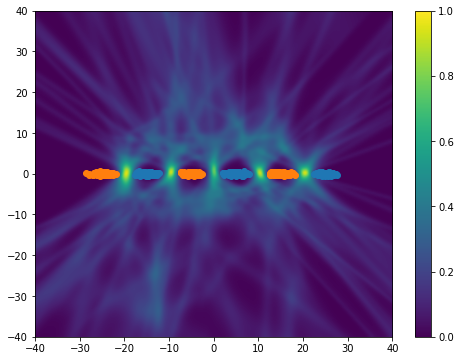}
         \caption{DE Aleatoric}
     \end{subfigure}
    \hfill
    \begin{subfigure}[b]{0.3\columnwidth}
         \centering
         \includegraphics[width=\columnwidth]{ims/toy-manifold/de_epistemic.png}
         \caption{DE Epistemic}
     \end{subfigure}
     \hfill
     \begin{subfigure}[b]{0.3\columnwidth}
         \centering
         \includegraphics[width=\columnwidth]{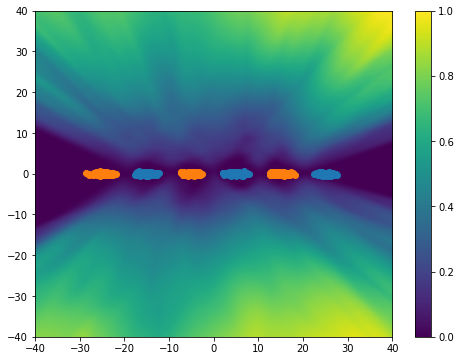}
         \caption{DE KL}
     \end{subfigure}
     \hfill
\end{figure}

\begin{figure}[h!]
     \begin{subfigure}[b]{0.3\columnwidth}
         \centering
         \includegraphics[width=\columnwidth]{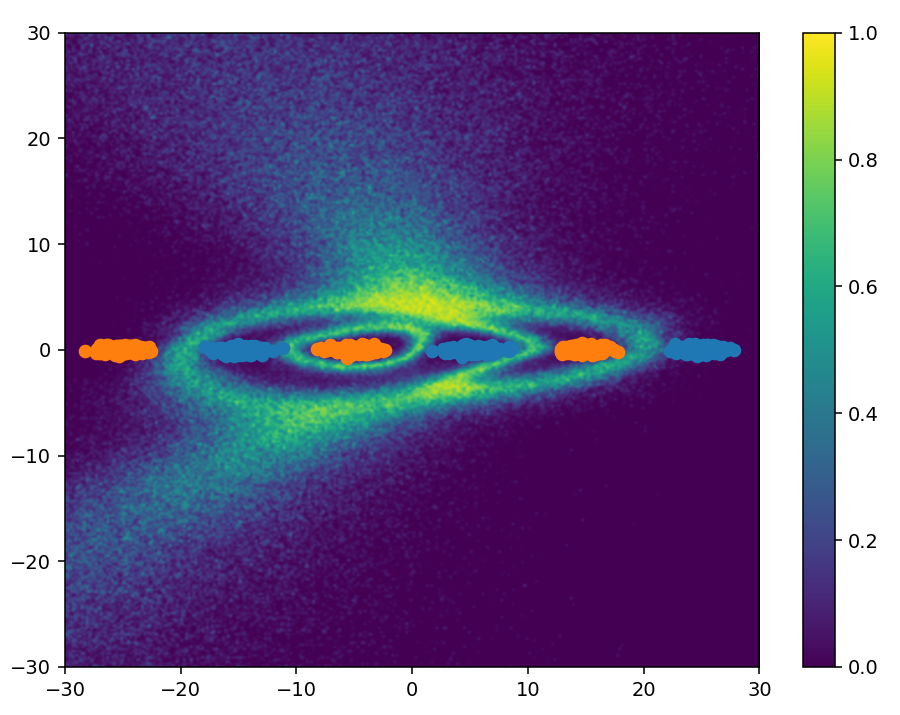}
         \caption{\tiny{MC Dropout Aleatoric}}
     \end{subfigure}
    \hfill
    \begin{subfigure}[b]{0.3\columnwidth}
         \centering
         \includegraphics[width=\columnwidth]{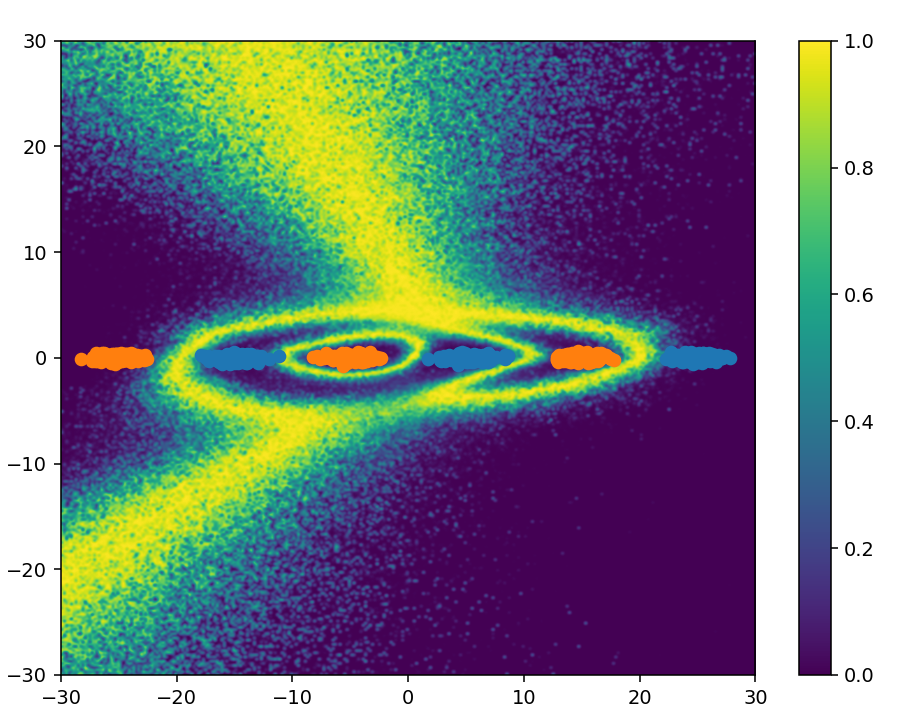}
         \caption{\tiny{MC Dropout Epistemic}}
     \end{subfigure}
     \hfill
     \begin{subfigure}[b]{0.3\columnwidth}
         \centering
         \includegraphics[width=\columnwidth]{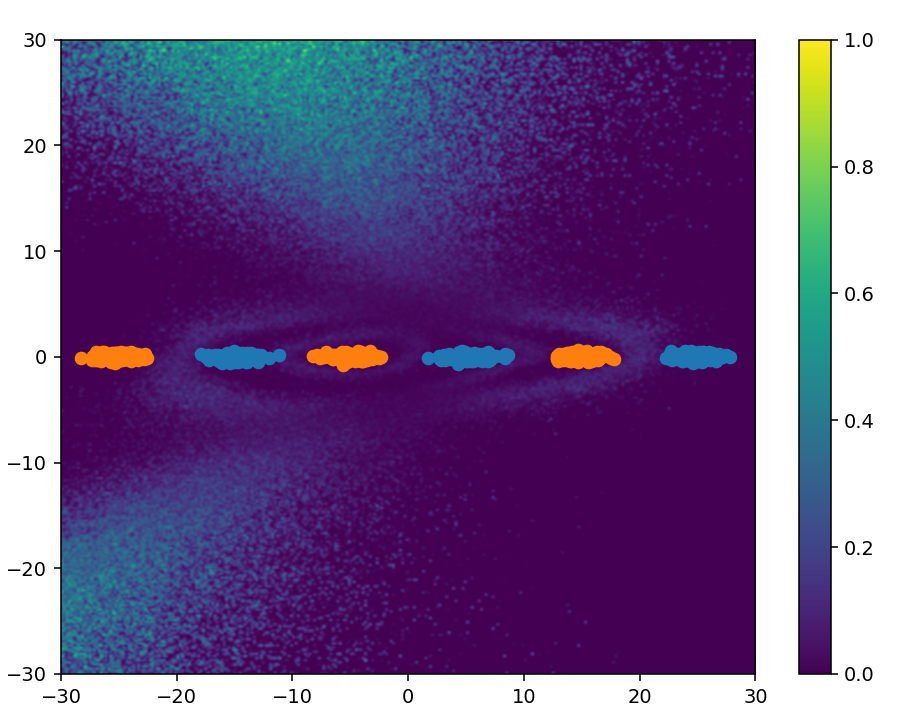}
         \caption{MC Dropout KL}
     \end{subfigure}
     \hfill
\end{figure}

\begin{figure}[h!]
     \begin{subfigure}[b]{0.3\columnwidth}
         \centering
         \includegraphics[width=\columnwidth]{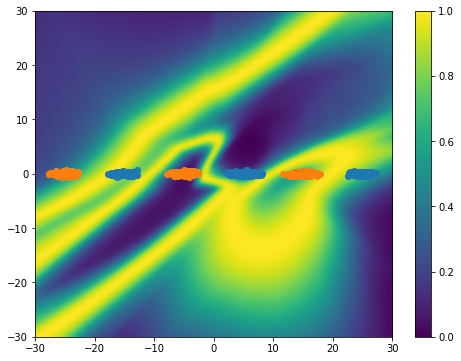}
         \caption{\tiny{SWAG Aleatoric}}
     \end{subfigure}
    \hfill
    \begin{subfigure}[b]{0.3\columnwidth}
         \centering
         \includegraphics[width=\columnwidth]{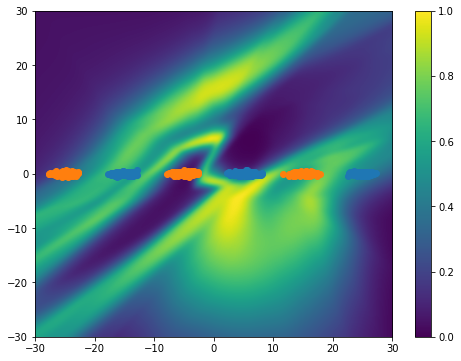}
         \caption{\tiny{SWAG Epistemic}}
     \end{subfigure}
     \hfill
     \begin{subfigure}[b]{0.3\columnwidth}
         \centering
         \includegraphics[width=\columnwidth]{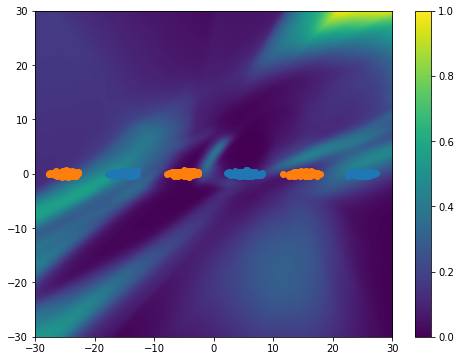}
         \caption{SWAG KL}
     \end{subfigure}
     \hfill
\end{figure}

\begin{figure}[h!]
     \begin{subfigure}[b]{0.3\columnwidth}
         \centering
         \includegraphics[width=\columnwidth]{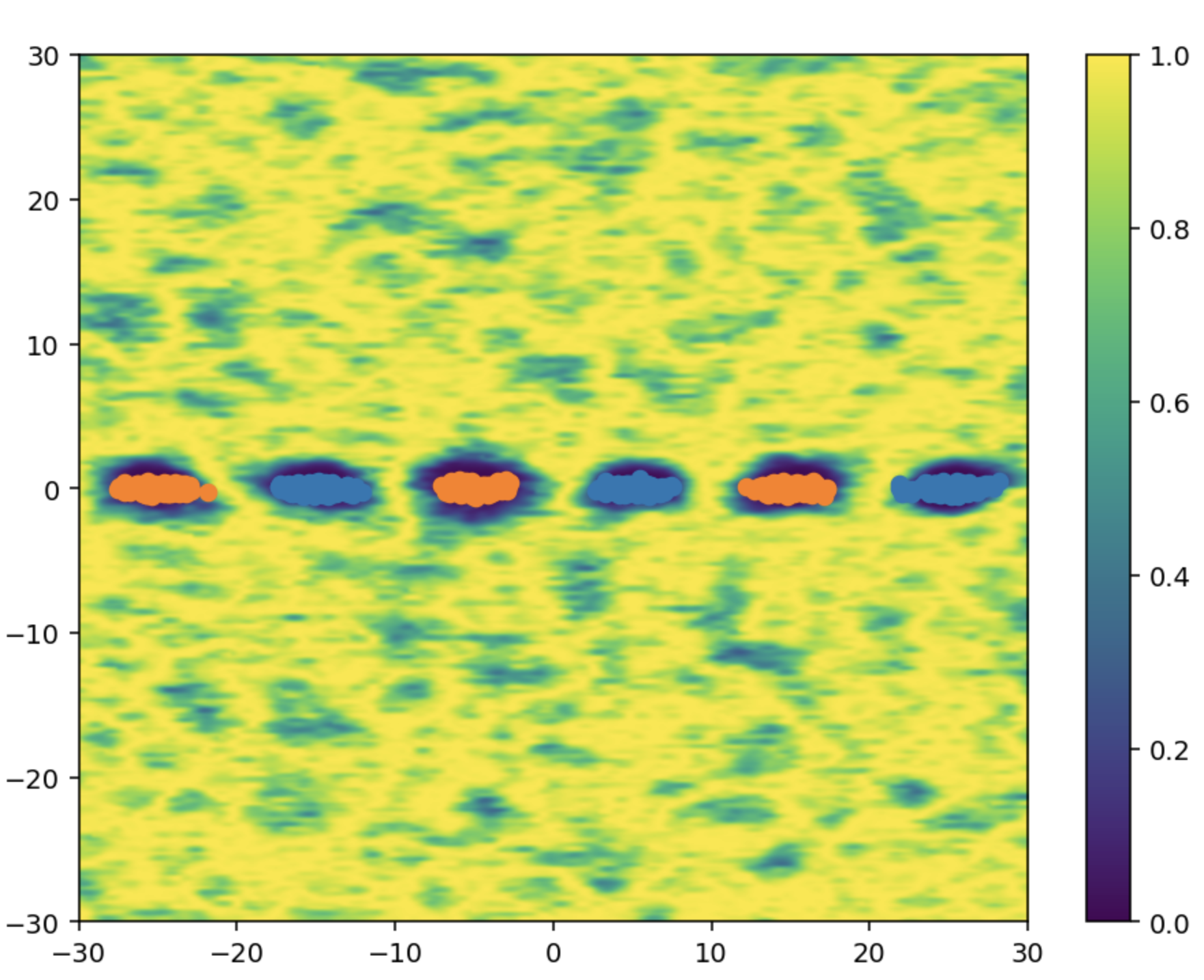}
         \caption{\tiny{SNGP Aleatoric}}
     \end{subfigure}
    \hfill
    \begin{subfigure}[b]{0.3\columnwidth}
         \centering
         \includegraphics[width=\columnwidth]{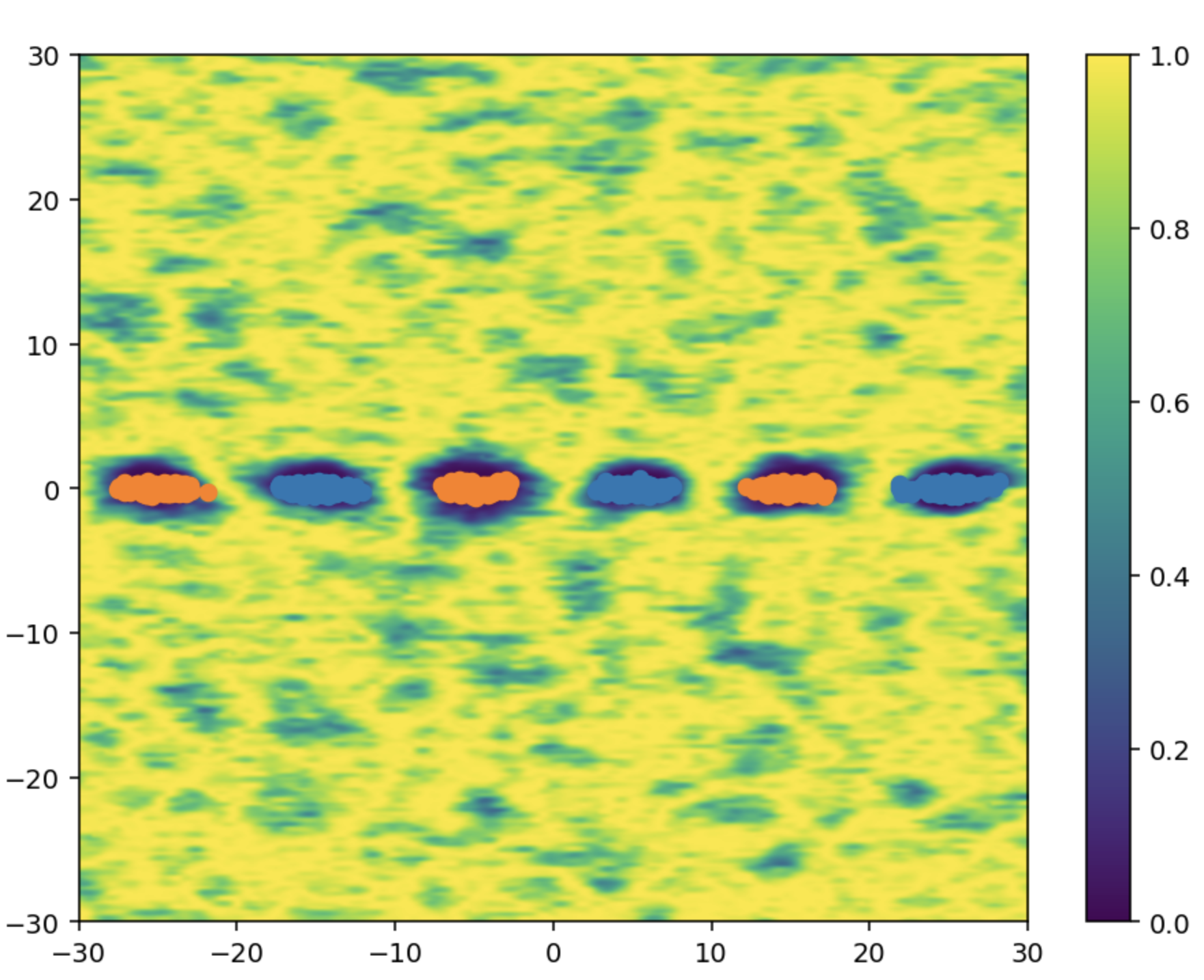}
         \caption{\tiny{SNGP Epistemic}}
     \end{subfigure}
     \hfill
     \begin{subfigure}[b]{0.3\columnwidth}
         \centering
         \includegraphics[width=\columnwidth]{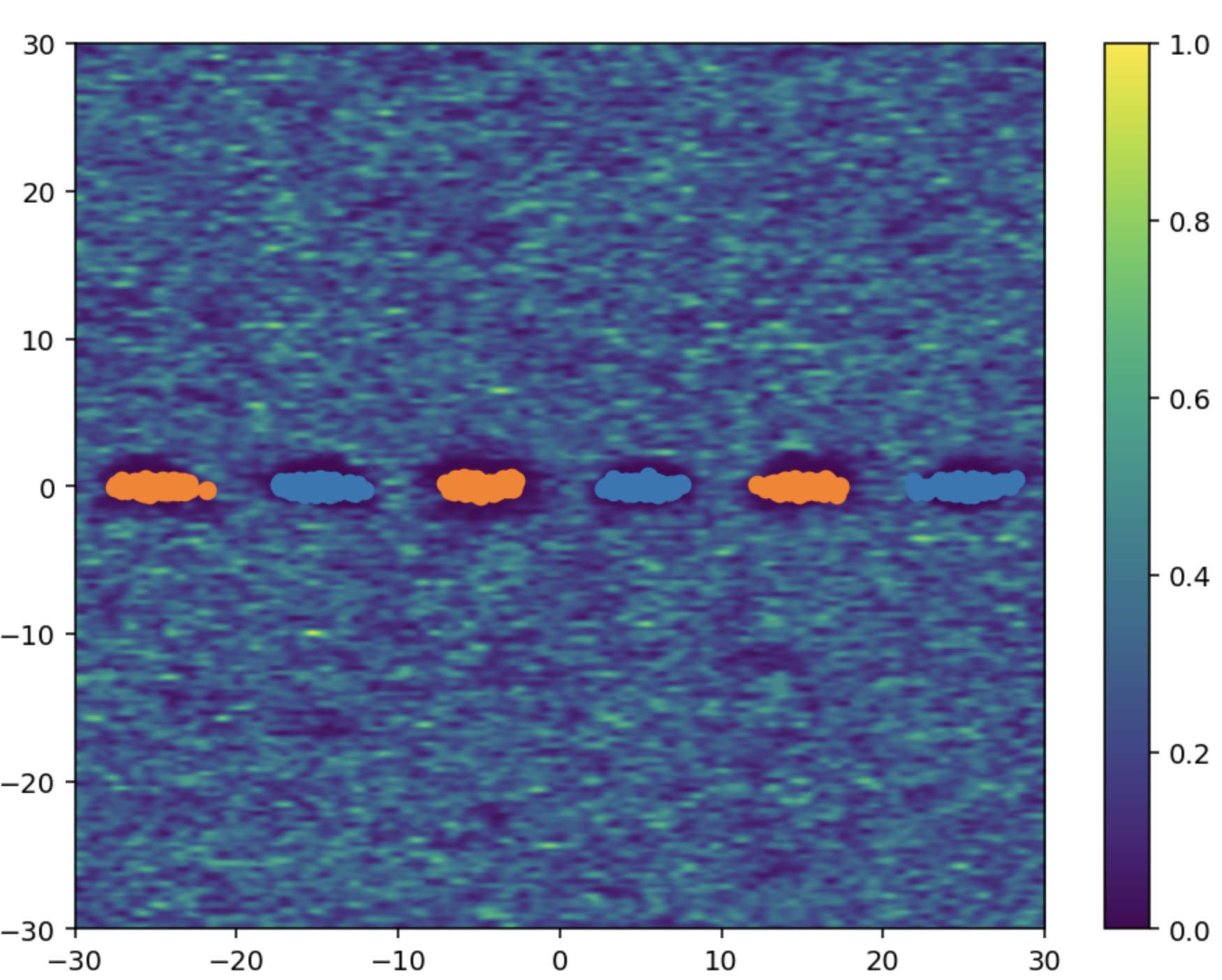}
         \caption{\tiny{SNGP KL}}
     \end{subfigure}
     \hfill
\end{figure}

\begin{figure}[h!]
     \begin{subfigure}[b]{0.4\columnwidth}
         \centering
         \includegraphics[width=\columnwidth]{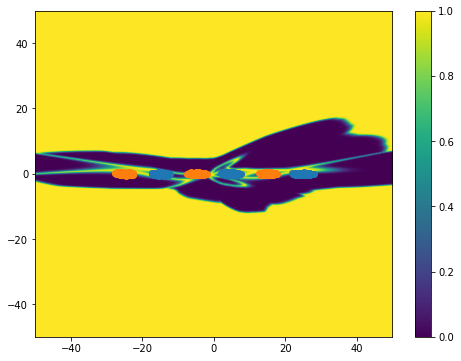}
         \caption{DUQ Aleatoric}
     \end{subfigure}
    \hfill
    \begin{subfigure}[b]{0.4\columnwidth}
         \centering
         \includegraphics[width=\columnwidth]{ims/toy-manifold/duq_epistemic.png}
         \caption{DUQ Epistemic}
     \end{subfigure}
\end{figure}

\end{document}